\documentclass{article}

\usepackage[main, final]{neurips_2026}
\usepackage{tabularx}

\usepackage[utf8]{inputenc} 
\usepackage[T1]{fontenc}    
\usepackage{hyperref}       
\usepackage{url}            
\usepackage{booktabs}       
\usepackage{amsfonts}       
\usepackage{nicefrac}       
\usepackage{microtype}      
\usepackage{xcolor}         
\usepackage{wrapfig}

\usepackage{makecell}
\usepackage{graphicx}
\usepackage{subcaption}

\usepackage{bbding}
\usepackage{iftex}

\ifPDFTeX

\else

\fi

\usepackage{amsmath}
\usepackage{amssymb}
\usepackage{mathtools}
\usepackage{amsthm}

\usepackage[capitalize,noabbrev]{cleveref}

\theoremstyle{plain}

\theoremstyle{definition}

\theoremstyle{remark}

\title{Beyond Proxy Metrics: MLLMs as Human Surrogates for Explainable AI Evaluation}

\author{%
  Eoin M.~Kenny\thanks{Corresponding author.} \\
  AI Research \\
  JP Morgan \\
  25 Bank Street, London, UK \\
  \texttt{eoin.kenny@jpmorgan.com} \\
  \And
  Salim I.~Amoukou \\
  AI Research \\
  JP Morgan \\
  25 Bank Street, London, UK \\
  \texttt{salim.ibrahimamoukou@jpmorgan.com} \\
  \And
  Tom Bewley \\
  Mistral AI \\
  London, UK \\
  \texttt{tom.bewley@mistral.ai}\thanks{Work done while at J.P. Morgan AI Research.} \\
}

\begin{document}

\maketitle

\begin{abstract}
Explainable AI (XAI) has expanded into a broad and impactful research area, but the fundamental question of how best to evaluate explanation techniques has proven frustratingly elusive. 
Methodologies typically revolve around either proxy metrics which tell us little about the utility of an explanation, or human evaluation which---while generally considered the gold standard---is expensive, difficult to scale, and challenging to reproduce. 
In this paper, we investigate if Multimodal Large Language Models (MLLMs) can act as surrogate human participants to automate the evaluation of XAI.
To do this, we first propose a ``purpose-grounded'' evaluation framework with metrics applicable to both MLLMs and human users alike. 
Then, we offer some conceptual insights as to why and when we can trust these models to automate this process.
Lastly, we conduct extensive empirical testing across five purpose-grounded areas such as model improvement and fourteen popular MLLMs. 
Results indicate that all models trend positively, but OpenAI and Qwen models generally show higher alignment compared to other variants, which would be the recommended choices in practice.
We expect our insights not to replace user testing, but rather to help XAI researchers complement their current evaluation methodologies and move the field away from heavy reliance on proxy metrics for measuring explanation utility.
\footnote{Code and raw data are released for reproducibility at: \url{https://github.com/EoinKenny/MLLM_Human_Surrogates}.}
\end{abstract}

\section{Introduction}

\begin{figure*}[!t]
\centering
  \includegraphics[width=\textwidth]{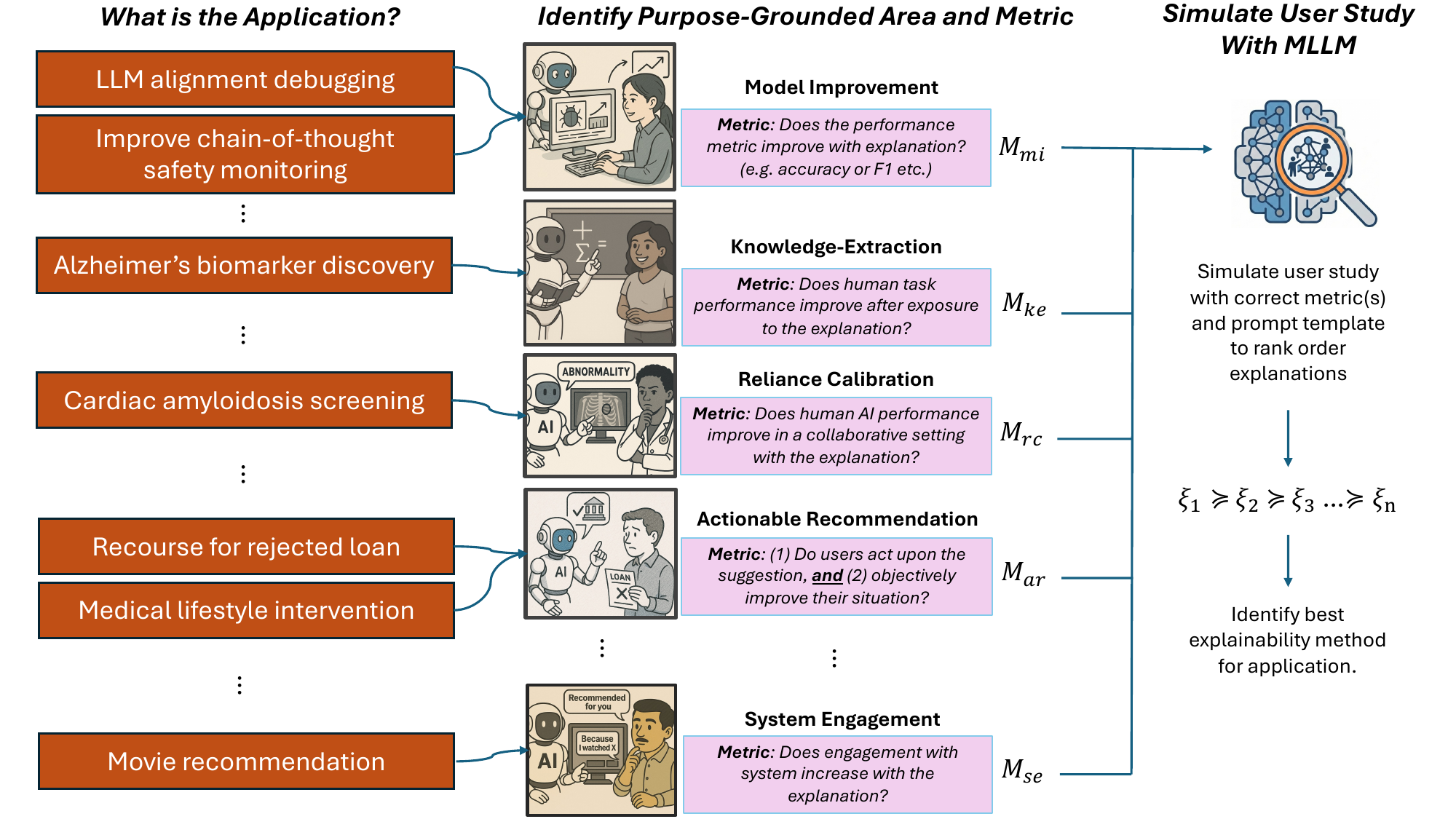}
  \caption{
  \textbf{Purpose-grounded evaluation framework}. Our framework for how XAI evaluation can be complemented using MLLMs as surrogate human users.
  We structure this evaluation around purpose-grounded domains in the literature because they contain clear evaluation metrics and are the areas XAI has proven most useful, here we show five of those.
  With both the purpose and metric to hand, researchers can simulate appropriate user testing to help identify if their method is more promising than current baselines or not.
  Importantly, note this framework can (and should) also be used with real humans in place of the MLLM, as we illustrate during this paper's experiments in Sections~\ref{sec:model_improvement}-\ref{sec:system_eng}.
  }
  \label{fig:title}
\end{figure*}

Explainable AI (XAI) has existed since at least the 1980s~\citep{munro1984review}, but its urgency has increased over the past decade as AI systems have become increasingly prevalent in high-stakes domains such as healthcare and criminal justice~\cite{barnett2024improving,bagaric2022solution}. 
In these contexts, understanding the rationale behind AI decisions is often as critical as the decisions themselves.
However, the proliferation of XAI methods has outpaced the development of meaningful evaluation methodologies.

Most studies and proposed evaluation metrics rely on proxy evaluations that rarely reflect an explanation's usefulness to end-users~\cite{freiesleben2023dear}.
For instance, counterfactual explanations frequently aim to minimize the $L_0$ norm, indicating the fewest feature changes required to alter an outcome~\cite{wachter2017counterfactual,keane2021if,kenny2023utility}. 
Yet, explanations optimized for such mathematical metrics often fail to offer actionable suggestions, or do not convince users to actually follow the advice~\cite{tominaga2024reassessing}, making their overall utility limited. 
This disconnect highlights a critical gap, proxy evaluations predominantly measure internal characteristics of explanations without sufficiently considering if this translates into achieving actual user goals or overall practical utility.
User studies can be run to verify true explanation usefulness if focused on outcome-driven metrics such as model improvement that serve a clear purpose, but they are notoriously challenging to design and reproduce~\cite{bhattacharya2024exmos,piorkowski2023aimee}.
As a result, researchers have begun exploring the usage of language models to automate this process, but the investigations so far have either not been XAI-specific~\cite{cui2024can} or narrowly focused~\cite{deevaluating}.

The idea of an explanation needing to serve a clear purpose has been discussed previously by several authors~\cite{hoffman2018metrics,doshi2017towards,freiesleben2023dear}.
The core tenet of this approach is that the development of XAI techniques should be guided by their intended downstream applications and practical utility, rather than by methodological novelty alone.
Several recent survey papers have begun converging on at least five clear areas for XAI in human-AI interaction, namely the fields of appropriate reliance~\cite{visser2023trust}, model improvement~\cite{gupta2024survey}, algorithmic recourse~\cite{karimi2022survey}, knowledge extraction~\cite{wu2023black}, and system engagement~\cite{zhang2020explainable}, so we use these as a starting point for our framework in Figure~\ref{fig:title}, but acknowledge more may exist.
Borrowing from the literature~\cite{doshi2017towards,freiesleben2023dear}, we refer to these as \textit{`purpose-grounded'} XAI areas.
We are aligned with all these authors conceptually, but interested in how MLLMs can help automate such evaluations.
Our contributions are:

\paragraph{Contributions}
(1) We propose a formal framework for purpose-grounded XAI evaluation in human-AI interaction, with precisely defined metrics across five areas (Section~\ref{sec:metrics}, Figure~\ref{fig:title}). 
(2) We provide a conceptual analysis of the conditions under which MLLMs can be trusted as surrogates for human responses to rank order explanation methods (Section~\ref{sec:theory}). 
(3) We contribute an empirical benchmark across five purpose-grounded domains and 14 MLLMs, identifying where rank replication robustly holds and systematic failure conditions (Section~\ref{sec:exp_setup}).

\section{Evaluation Metrics}
\label{sec:metrics}
In order to enable potential practitioners, we formally define all evaluation metrics for the usage of MLLMs---and indeed human participants---in our framework.
Specifically, we examine five purpose-grounded areas for XAI defined within a human-computer interaction setting, where a human stakeholder $H$ is receiving an explanation $\xi$ from any source, while performing some activity.
The explanation $\xi$ can be of any form (e.g., local, global, counterfactual, factual, and combinations thereof).
Furthermore, we assume that each $(H, AI)$ pairing is sampled from a target \textit{population} $\rho$, which could contain humans with different levels of expertise, and AI systems with different architectures, objectives and performance levels.\footnote{Special cases of this formulation can involve multiple humans paired with the same AI, or vice versa.}
Given such a population, each purpose $p$ is associated with an explanation evaluation metric $M_p$, defined as the expected task-level performance induced by interaction with the explanation. 
Concretely, all evaluation metrics considered in this paper follow the generic pattern
\begin{equation}
M_p(\xi)
=
\mathbb{E}_{(H, AI)\sim\rho}
\bigl[\phi(\cdot^+)\bigr],
\quad \text{where}\quad
\cdot^+ = \text{Interaction}(\cdot \mid \cdot, \xi)
\label{eq:generic_pattern}
\end{equation}
where $\phi$ is a scalar criterion appropriate to purpose $p$, and
$\cdot$ denotes the entity being evaluated (e.g., the AI system, the
human, or the human--AI team), and $\text{Interaction}(\cdot \mid \cdot,
\xi)$ captures the way in which interaction with explanation $\xi$
induces an updated state $\cdot^+$. 
We now instantiate this into our five areas illustrated in Figure~\ref{fig:title}.

\paragraph{Model Improvement (MI)}
Here, the objective is to determine if an explanation $\xi$ aids an engineer or developer $H$ in improving the system $AI$ according to some scalar performance criterion $\phi$ (e.g., accuracy, robustness, fairness), for which higher values are better. 
Let $AI^+$ denote the version of $AI$ improved by $H$ using explanation $\xi$. 
Aggregating over the population $\rho$, the evaluation metric $M_{mi}$ is
\begin{equation}
M_{mi}(\xi)
=
\mathbb{E}_{(H, AI)\sim\rho}
\bigl[\phi(AI^+)\bigr],
\quad \text{where}\quad
AI^+ = \text{Improve}(AI \mid H, \xi)
\end{equation}
where the conditioning ``$|\ H,\xi$'' indicates that the way in which $AI$ is improved depends on how $H$ interacts with the provided explanation.
For example, an engineer ($H$) could use feature attribution explanations ($\xi$) to debug and retrain a classifier model ($AI$), with the aim of obtaining an improved model ($AI^+$) with better adversarial robustness ($\phi$).

\paragraph{Knowledge Extraction (KE)}
This purpose assesses if an explanation $\xi$ from system $AI$ effectively teaches a concept or skill to human $H$, which can be something known by domain experts, or new scientific knowledge. 
Success is measured by $H$'s ability to apply the learned knowledge independently on a downstream task, where performance is again measured by a scalar criterion $\phi$ (e.g., score on a post-teaching quiz). 
The evaluation metric $M_{ke}$ is
\begin{equation}
M_{ke}(\xi)
=
\mathbb{E}_{(H, AI)\sim\rho}
\bigl[\phi(H^+)\bigr],
\quad \text{where}\quad
H^+ = \text{Teach}(H \mid AI, \xi)
\end{equation}
where the conditioning ``$|\ AI,\xi$'' indicates that the provided explanation is used in conjunction with any other outputs from $AI$ to facilitate the teaching of $H$.
For example, this could measure the accuracy ($\phi$) of a medical student ($H$) when diagnosing, after being taught by an AI using rule-based explanations ($\xi$).

\paragraph{Reliance Calibration (RC)}
In this setting, $H$ and $AI$ are jointly working to perform a task in a human-AI team, denoted by $H\hspace{-0.07cm}AI$.
$H$ is assumed to have the ultimate decision-making authority about which actions are taken, and thus must decide when to rely on (and conversely, when to disregard) the outputs of $AI$.
The purpose of an explanation $\xi$ in this setting is to assist $H$ in adapting their reliance strategy, such that the complementary performance of the human-AI team (once again defined by a scalar criterion $\phi$) is improved.
The evaluation metric $M_{rc}$ is therefore
\begin{equation}
M_{rc}(\xi)
=
\mathbb{E}_{(H, AI)\sim\rho}
\bigl[\phi(H\hspace{-0.07cm}AI^+)\bigr],
\quad \text{where}\quad
H\hspace{-0.07cm}AI^+
=
\text{Adapt}(H\hspace{-0.07cm}AI \mid \xi)
\end{equation}
where the conditioning ``$|\ \xi$'' indicates that the adapted behavior of the human-AI team depends on the given explanation.
For example, this metric could measure the accuracy of predicting phantom braking ($\phi$) by a safety driver ($H$) working with a self-driving car ($AI$), where the car's chosen trajectories are augmented by concept-based explanations ($\xi$)~\cite{kenny2026explainable}.

\paragraph{Actionable Recommendation (AR)}
This purpose evaluates if an explanation $\xi$ provides effective and adoptable actions for a user $H$ that is subject to decisions made by $AI$. 
A canonical example of this case is automated credit scoring and loan approval, for which a rejected applicant may wish to take recourse actions to become accepted in future.
Effectiveness depends on two factors: whether or not the user adopts the recommended action, denoted by the binary variable `$\text{Follows}(H\ |\ \xi)$', and the positive benefit derived from it if they do, `$\text{Benefit}(H,AI\ |\ \xi)$'.
Note that the concept of benefit can encompass direct benefits to $H$ (e.g., the money they receive as their loan) but also to other stakeholders (e.g., the interest that the bank stands to gain during repayment).
The evaluation metric $M_{ar}$ takes the product of these two factors for each $(H, AI)$ pair, and therefore gives the expected benefit weighted by the likelihood of adoption across the population:
\begin{equation}
M_{ar}(\xi)
=
\mathbb{E}_{(H, AI)\sim\rho}
\bigl[
\text{Follows}(H \mid \xi)
\times
\text{Benefit}(H, AI \mid \xi)
\bigr].
\label{eq:recourse}
\end{equation}
For $M_{ar}$ to be high, reflecting greater overall utility, both components must contribute positively. 
This signifies that the explanation leads to recommendations that are both followed and beneficial.
For instance, robust explanations that are valid over time might have greater potential benefits~\cite{dominguez2022adversarial}, while clear, simple and persuasive explanations might have higher rates of being followed~\cite{tominaga2024reassessing}.

\paragraph{System Engagement (SE)}
This purpose evaluates whether an explanation $\xi$ enhances the engagement of human $H$ with system $AI$, where this system is typically one that recommends content for entertainment or education. 
Engagement is measured by a suitable metric (e.g., interaction time or depth). 
The evaluation metric $M_{se}$ is:
\begin{equation}
M_{se}(\xi)
=
\mathbb{E}_{(H, AI)\sim\rho}
\bigl[\text{Engagement}(H, AI \mid \xi)\bigr].
\end{equation}
An example is measuring how frequently movie streaming platform users ($H$) follow the suggestions of a recommender system ($AI$) when textual explanations ($\xi$) for those recommendations are provided.

\subsection{Ranking explanations}
\label{sec:ranking_explanations}
For any given explanation $\xi$, each of these purpose-grounded metrics only gains meaning when compared to alternatives.
In general, studies should consider a set of $n$ candidate explanations (including a no-explanation control), and construct a ranking of this set by comparing their metric values over a population $\rho$ of experimental subjects and AI systems.
For a given purpose $p$, the ranking is defined as follows:
\begin{equation}
\xi_i \succeq_p \xi_j
\Leftrightarrow
M_p(\xi_i) \geq M_p(\xi_j),
\qquad
\xi_1 \succeq_p \xi_2 \succeq_p \cdots \succeq_p \xi_n.
\label{eq:ranking}
\end{equation}
When determining the best explanation for $p$, it is this ranking that is of particular importance, not the raw metric scores themselves.
This insight is inspired by recent research replicating human studies with LLMs and peer review~\cite{cui2024can,deevaluating,ramdas2026aistats}, illustrating that main effects generally replicate, but not the subtleties regarding actual means and distributions.

\section{Conceptual Analysis of the Ranking Phenomenon}
\label{sec:theory}
The framework in Section~\ref{sec:metrics} compares rankings rather than requiring agreement in raw metric values (Equation~\ref{eq:ranking}). 
This distinction is motivated by research comparing human and MLLM responses in experiments and peer review~\cite{cui2024can,deevaluating,ramdas2026aistats}.
We sketch an argument for how rankings may be preserved despite differences in response distributions, and where this may break down.
Our argument uses two working assumptions: (1) the MLLM has learned response patterns relevant to the task, allowing its responses to track aspects of human responses imperfectly; and (2) the comparison involves structurally distinct explanation types, such as counterfactual explanations and control.

\subsection{Human heterogeneity and MLLM sampling variability}
Consider a study population $\rho$ in which each individual $h$ has latent attributes $\eta_h$ (e.g., expertise, attention).
Given an explanation $\xi$, individual $h$ produces a response $y$ drawn from $P(y \mid \xi, \eta_h)$. 
The full population response distribution is $P_H(y \mid \xi) = \int P(y \mid \xi, \eta_h)\, d\rho(\eta_h)$, with variance $\sigma_H^2(\xi) := \text{Var}_{P_H(y\mid\xi)}(y)$. 
By the law of total variance,
\begin{equation}
\sigma_H^2(\xi) =
\underbrace{\mathbb{E}_{\eta_h}[\text{Var}(y \mid \xi, \eta_h)]}_{\text{individual stochasticity}}
+
\underbrace{\text{Var}_{\eta_h}[\mathbb{E}(y \mid \xi, \eta_h)]}_{\text{population heterogeneity}}.
\label{eq:variance_decomp}
\end{equation}
The first term captures variation within individuals and the second systematic differences across people.
For a fixed MLLM and prompt, let $p_\theta(y \mid \xi; T)$ denote its response distribution at temperature $T$, with variance $\sigma_M^2(\xi) := \text{Var}_{p_\theta(y\mid\xi;T)}(y)$.
Repeated sampling captures model-output variability, but does not establish that human population heterogeneity has been reproduced.

\subsection{Why ordering may survive compression}
\label{sec:ordering}
Let $\mu_H(\xi)$ and $\mu_M(\xi)$ denote the human and MLLM means for explanation $\xi$. 
These need not match, and we define
\begin{equation}
\mu_M(\xi) = \mu_H(\xi) + \epsilon(\xi),
\end{equation}
where $\epsilon(\xi)$ is a condition-specific calibration error.
An analogy is how two miscalibrated thermometers can still order temperatures correctly when the difference between their errors is smaller than the temperature gap.
Similarly, if MLLM calibration errors share a common additive component across conditions, that component cancels under pairwise comparison; the remaining difference determines whether the ordering is preserved.

Formally, fix two conditions with $\mu_H(\xi_i) > \mu_H(\xi_j)$. 
A sufficient condition for the MLLM to preserve this ordering is
\begin{equation}
|\epsilon(\xi_i) - \epsilon(\xi_j)|
<
|\mu_H(\xi_i) - \mu_H(\xi_j)|,
\label{eq:ranking_condition}
\end{equation}
i.e., the difference in calibration errors is smaller in magnitude than the human mean gap.
This follows because the MLLM mean difference equals the human mean difference plus $\epsilon(\xi_i)-\epsilon(\xi_j)$.
The condition is sufficient but not necessary: when the difference in calibration errors aligns in sign with the human mean gap, it amplifies rather than reverses the ordering, even when Equation~\ref{eq:ranking_condition} fails.
Shared response patterns may produce similar calibration errors across conditions, helping preserve their ordering. 
When the difference in calibration errors is large relative to the human mean gap, Equation~\ref{eq:ranking_condition} may fail, allowing but not requiring an ordering reversal.
This illustrates a possible failure mechanism for MLLM surrogates.

\section{Experiments}
\label{sec:exp_setup}

We evaluate MLLMs using adaptations of five human studies spanning the purpose-grounded areas in Figure~\ref{fig:title}.
The benchmark includes 14 models: nine API models (GPT-4o, GPT-4o mini, o4-mini, GPT-5, GPT-5 mini, GPT-5.5, Claude 4.5 Sonnet, Claude 4.6 Sonnet, and Claude 4.8 Opus) and five open-weight models (Gemma 3 27B, Gemma 4 31B, Qwen3-VL 32B, Qwen3.5 122B, and Mistral Small 3.2).
Exact model identifiers and inference settings are documented in Appendix~\ref{app:model_ids}.
The point is not only to demonstrate examples of how our framework fits in real setups, but also how well MLLMs can replicate the results of real human testing.
These specific studies were chosen because they illustrate well the five purpose-grounded areas, but they are also methodologically sound in their design (e.g., reasonable samples and control conditions).
We reached out to the authors when needed for clarifications and for permission to use their figures.
Building upon the discoveries of similar research~\cite{cui2024can}, all our experiments used a temperature of 1.0 when prompting the MLLM.

We report full-data Spearman's $\rho$ between MLLM condition means and fixed human reference rankings, where $\rho=1$ indicates identical rankings, $\rho=0$ no monotonic association, and $\rho=-1$ reversed rankings.
We estimate 95\% percentile intervals using 10,000 paired bootstrap resamples with replacement, each at the observed sample size.
Strong positive correlations with narrow intervals entirely above zero indicate general trending agreement with the human reference that is stable under resampling of the collected MLLM responses.

To translate explanation information into prompt format for the MLLMs, we took a variety of approaches depending on the study.
The reliance calibration and knowledge extraction studies used a pre-test phase in which participants were taught about the domain, so we included examples in the prompt to simulate this.
The model improvement study used a GUI and repeated interactions, so we included an image of the data distribution shown to users in the real study to simulate this cognitive load, and accumulated all past interactions in the prompt summarised for each future interaction.
Feature attribution information was written as text with two floating point numbers.
Full details for all studies are in the appendix.

\subsection{Model Improvement}
\label{sec:model_improvement}

\begin{figure*}[t!]
\centering
  \includegraphics[width=\textwidth]{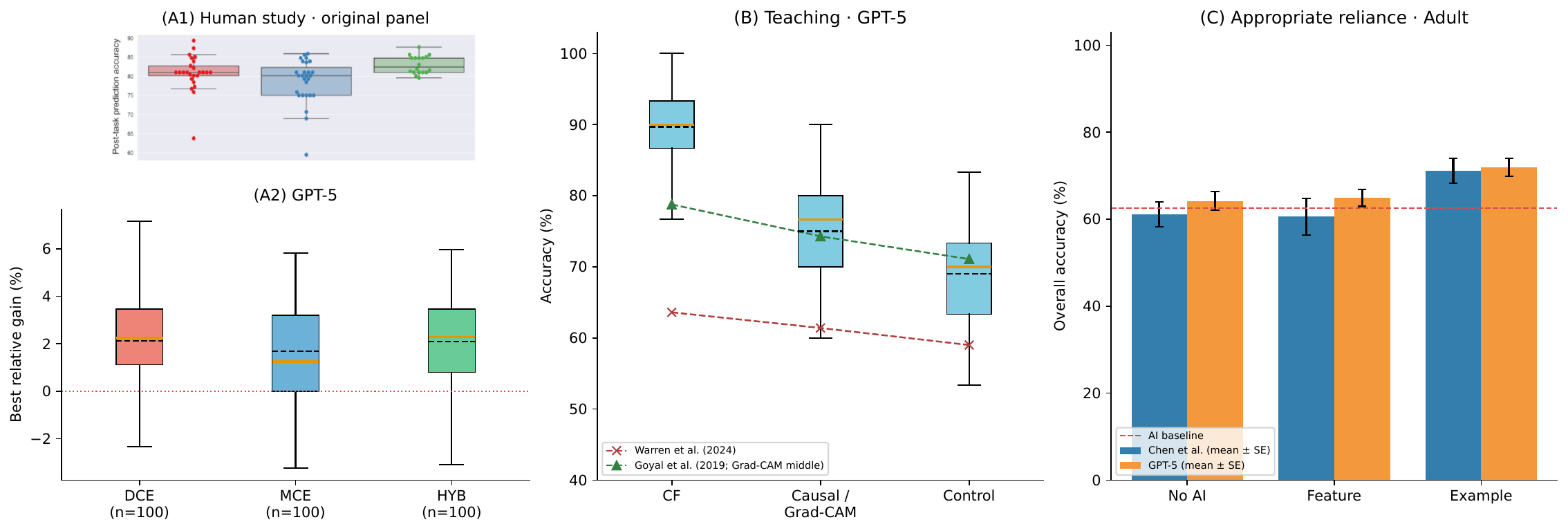}
    \caption{
    \textbf{Human references and GPT-5 results.}
    (A1) Model-improvement panel from Bhattacharya et al.~\cite{bhattacharya2024exmos}.
    (A2) GPT-5 relative accuracy improvement over each run’s baseline, using the best successful proposal among five attempts; boxes summarize 100 runs per condition.
    (B) GPT-5 synthetic teaching accuracy across 10,000 paired bootstrap resamples of 30 questions, with human reference means from Warren et al.~\cite{warren2024categorical} and Goyal et al.~\cite{goyal2019counterfactual}. 
    (C) Adult classification accuracy from Chen et al.~\cite{chen2023understanding} and GPT-5. Human bars show the published means and standard errors; GPT-5 bars show means and standard errors across 26 dataset-seed blocks. 
    }
  \label{fig:results1}
\end{figure*}

This area concerns improving model properties such as accuracy, fairness, safety, robustness, and alignment~\cite{gupta2024survey,weber2023beyond}, with researchers free to define the metric of improvement. 
Bhattacharya et al.~\cite{bhattacharya2024exmos} evaluated how explanation type supports model accuracy as the final metric. 
In a study with 70 healthcare professionals, participants used the EXMOS interactive system with one of three explanation dashboards: a data-centric version (DCE) visualising dataset characteristics and quality issues, a model-centric version (MCE) showing SHAP-based feature importance and surrogate decision rules, and a hybrid (HYB) combining both. 
Participants modified the training data based on explanations and retrained the model; HYB significantly outperformed DCE and MCE in post-retraining accuracy (Figure~\ref{fig:results1}(A1)). 
In our framework, $H$ are physicians, $AI$ is a random forest classifier, $\xi_1, \xi_2, \xi_3$ are DCE, MCE, HYB, and $\phi(\cdot)$ is the post-retraining accuracy delta.

Each MLLM was given five iterative attempts to improve accuracy, receiving a screenshot of the data distribution (Figure~\ref{fig:model_improv_distributions}) together with feature-importance scores, decision rules, or dataset statistics according to its assigned condition. 
Each iteration contained information on all prior attempts, and we recorded the best configuration's delta. 
We repeated this over 100 seeds with different data splits, yielding a per-condition mean delta-accuracy and an empirical condition ranking for each MLLM.
To test whether this ranking matches the human ground truth (HYB $>$ DCE $>$ MCE) we computed Spearman's $\rho$. 
Precision was quantified by 95\% confidence intervals from bootstrap resamples.

Table~\ref{tab:results_all} shows five bootstrap intervals entirely above zero. 
Agreement nevertheless remains uncertain for several models, as discussed in Section~\ref{sec:theory}, rankings can be fragile when differences between human condition means are small or explanation types overlap, which is the case for the original human study. 
In addition, our post-hoc analysis (Appendix~\ref{app:mi-case-analysis}) identifies difficulties interpreting supplied information and using iterative feedback, alongside coarse feedback-set score resolution. 
These findings highlight challenges in this task, however 12/14 models did trend positively.

\subsection{Knowledge Extraction}

This area concerns the ability of explanations to teach humans new knowledge, either previously known~\cite{goyal2019counterfactual} or entirely novel scientific insights~\cite{schut2023bridging}. 
We focus on Warren et al.~\cite{warren2024categorical}, who compared counterfactual, causal, and control explanations for teaching humans about blood alcohol content. 
Participants underwent a training phase using one of the three explanation types before a testing phase requiring predictions on new data. 
The counterfactual group achieved the best accuracy (63.6\%), followed by causal (61.4\%) and control [59.0\%; Figure~\ref{fig:results1}(B)], converging with prior findings~\cite{goyal2019counterfactual}. 
In our framework, $H$ are non-experts, $AI$ is the Widmark equation, and $\xi_1, \xi_2, \xi_3$ are control, causal, and counterfactual; $\phi(\cdot)$ is accuracy on test data.

\begin{figure*}[t!]
\centering
  \includegraphics[width=\textwidth]{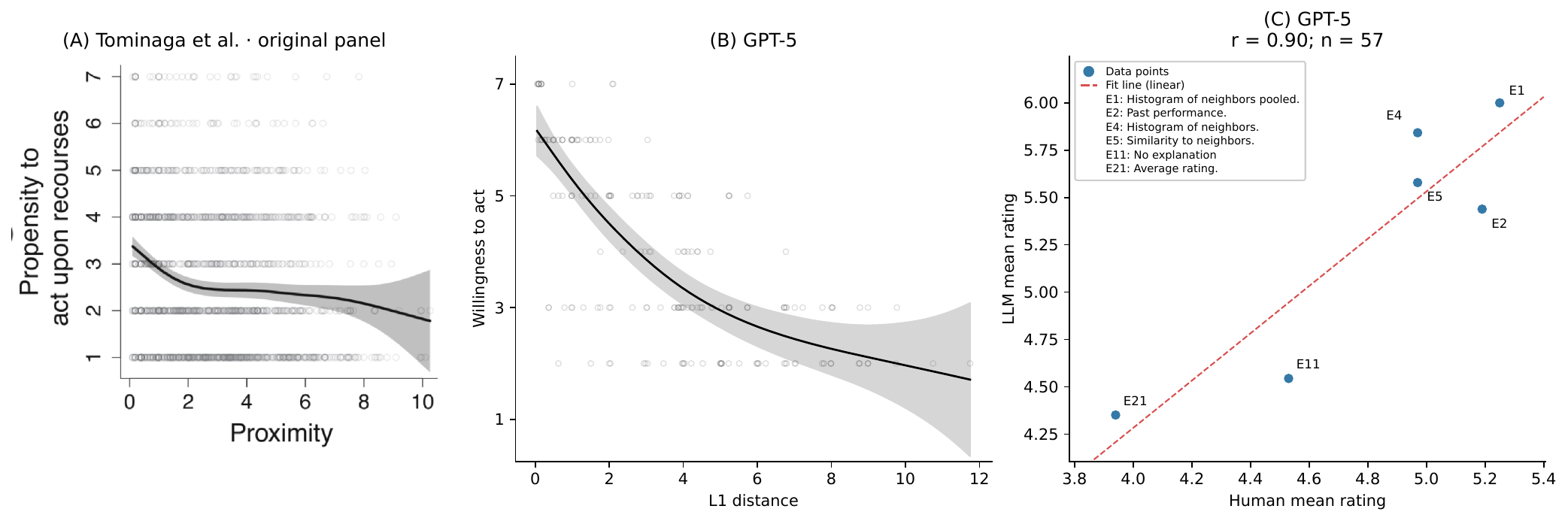}
    \caption{
    \textbf{Human references and GPT-5 results.}
    (A) Original willingness-to-act panel from Tominaga et al.~\cite{tominaga2024reassessing}.
    (B) GPT-5's equivalent ratings versus $L_1$ distance for 200 recourse plans sharing 39 applicants. 
    The curve is a fitted population-mean GAMM trend, with an approximate pointwise 95\% mean interval. 
    (C) Mean viewing-intention ratings for six explanation conditions from Herlocker et al.~\cite{herlocker2000explaining}. 
    Pearson's $r=0.90$ is computed across the six condition means. 
    The dashed line is a descriptive linear fit. 
    }  \label{fig:results2}
\end{figure*}

Because the Widmark equation is widely documented and likely present in MLLM pre-training data~\cite{widmark1981principles}, we constructed a synthetic dataset matching the original's feature count, categorical structure, and value ranges, but with anonymised feature names (e.g., \texttt{x1}, \texttt{x2}, \dots) and random causal relationships to increase task difficulty (Appendix~\ref{app:hmt_setup}).
All conditions saw the same training examples and labels; the counterfactual group received the mutations needed to flip each prediction, and the causal group additionally received feature-importance scores derived as the same feature used in the counterfactual perturbation. 
Each MLLM answered 300 test questions per condition.

Table~\ref{tab:results_all} shows positive teaching correlations for 13 of the 14 models, with 11 bootstrap intervals entirely above zero. 
Eight models reproduce the human reference ordering exactly in their point estimates ($\rho=1$). 
Thus, most models show stable positive agreement in this synthetic adaptation, although uncertainty remains model-dependent.

\subsection{Reliance Calibration}
\label{sec:reliance_calibration}

This pillar concerns whether users appropriately rely on an AI system to improve some aspect of performance. 
We focus on Chen et al.~\cite{chen2023understanding}, who measured whether explanations could calibrate reliance to improve classification performance, testing three conditions (control, saliency, and example-based) across the same dataset in tabular~\cite{adult_2} and NLP~\cite{bias_in_bios} domains.
Example-based explanations were the only condition to improve overall performance in either study. 
In our framework, $H$ are non-experts, $AI$ is a random forest (tabular) or bag-of-words classifier (NLP), $\xi_1, \xi_2, \xi_3$ are no-AI, saliency, and example-based, and $\phi(\cdot)$ is accuracy on novel test instances.

We adapted the Adult and BIOS tasks using reconstructed study materials. 
Each MLLM received five fixed labelled demonstrations and answered 16 cases in each of 26 dataset-seed blocks under three matched conditions: no AI advice, AI advice with  saliency~\cite{ribeiro2016should}, and AI advice with two nearest-neighbour examples. 
We compare full-data condition means with the human reference orderings: nearest neighbours $>$ control $>$ saliency for Adult, and nearest neighbours $>$ saliency $>$ control for BIOS.
The latter showed high unaided MLLM accuracy, suggesting limited headroom and a mismatch with the human reliance setting. We therefore report Adult as the primary reliance analysis and retain BIOS in Appendix~\ref{app:replication_details} for transparency.

All 14 models have positive Adult point estimates, with eight intervals entirely positive.
Frontier APIs such as GPT-5 mini, Claude Opus 4.8, and Claude Sonnet 4.6 perform best in general, but other open source models such as Mistral Small 3.2 also do equally well.

\begin{table*}[t]
\centering
\caption{Full-data Spearman's $\rho$ estimates with 95\% percentile intervals from 10,000 bootstrap resamples at each dataset's observed sample size. 
Bold entries have a nonnegative lower interval bound (including zero). 
Overall weights five studies equally. 
Recourse uses negative $L_1$ distance versus willingness and conditions on the observed applicant pool. 
Intervals condition on fixed human references.}
\label{tab:results_all}
\resizebox{\textwidth}{!}{\begin{tabular}{lcccccc}\toprule
Model & Teaching & Reliance & Engagement & Model improvement & Recourse & Overall \\\midrule
GPT-5 mini & \textbf{1.00 [0.50, 1.00]} & \textbf{1.00 [0.50, 1.00]} & \textbf{0.90 [0.84, 0.90]} & \textbf{1.00 [0.50, 1.00]} & \textbf{0.50 [0.37, 0.61]} & \textbf{0.88 [0.66, 0.90]} \\
Qwen3.5 122B & \textbf{1.00 [0.50, 1.00]} & 0.87 [-0.50, 1.00] & \textbf{0.81 [0.81, 0.81]} & 1.00 [-1.00, 1.00] & \textbf{0.50 [0.38, 0.60]} & \textbf{0.83 [0.26, 0.87]} \\
Claude 4.5 Sonnet & \textbf{1.00 [0.00, 1.00]} & 0.50 [-0.50, 1.00] & \textbf{0.75 [0.75, 0.81]} & 1.00 [-1.00, 1.00] & \textbf{0.52 [0.41, 0.62]} & \textbf{0.76 [0.28, 0.86]} \\
GPT-5.5 & \textbf{1.00 [0.50, 1.00]} & 0.50 [-0.50, 1.00] & \textbf{0.84 [0.78, 0.84]} & 0.50 [-1.00, 1.00] & \textbf{0.73 [0.64, 0.79]} & \textbf{0.71 [0.41, 0.91]} \\
GPT-5 & \textbf{1.00 [1.00, 1.00]} & \textbf{0.50 [0.50, 1.00]} & \textbf{0.81 [0.75, 0.90]} & \textbf{0.50 [0.50, 1.00]} & \textbf{0.74 [0.65, 0.80]} & \textbf{0.71 [0.69, 0.92]} \\
GPT-4o mini & 0.50 [-0.50, 0.50] & \textbf{0.50 [0.50, 1.00]} & \textbf{0.87 [0.72, 0.93]} & \textbf{1.00 [0.50, 1.00]} & \textbf{0.29 [0.14, 0.43]} & \textbf{0.63 [0.40, 0.75]} \\
GPT-4o & \textbf{1.00 [1.00, 1.00]} & \textbf{0.50 [0.50, 0.50]} & \textbf{0.64 [0.55, 0.64]} & 0.50 [-0.50, 1.00] & \textbf{0.48 [0.37, 0.58]} & \textbf{0.62 [0.40, 0.73]} \\
Qwen3-VL 32B & \textbf{0.50 [0.50, 1.00]} & 0.50 [-0.50, 1.00] & \textbf{0.81 [0.64, 0.81]} & \textbf{1.00 [0.50, 1.00]} & \textbf{0.30 [0.17, 0.41]} & \textbf{0.62 [0.32, 0.74]} \\
o4-mini & \textbf{0.50 [0.50, 1.00]} & \textbf{0.50 [0.50, 1.00]} & \textbf{0.90 [0.84, 0.90]} & \textbf{1.00 [0.50, 1.00]} & \textbf{0.15 [0.001, 0.30]} & \textbf{0.61 [0.50, 0.81]} \\
Claude 4.8 Opus & \textbf{1.00 [1.00, 1.00]} & \textbf{1.00 [0.50, 1.00]} & \textbf{0.79 [0.75, 0.81]} & -0.50 [-1.00, 0.51] & \textbf{0.74 [0.66, 0.81]} & \textbf{0.61 [0.40, 0.82]} \\
Claude 4.6 Sonnet & \textbf{1.00 [0.50, 1.00]} & \textbf{1.00 [0.50, 1.00]} & \textbf{0.75 [0.75, 0.81]} & -0.50 [-1.00, 0.50] & \textbf{0.76 [0.68, 0.82]} & \textbf{0.60 [0.41, 0.82]} \\
Gemma 4 31B & \textbf{0.50 [0.50, 0.50]} & 0.50 [-1.00, 1.00] & \textbf{0.75 [0.75, 0.75]} & 0.50 [-0.50, 0.50] & \textbf{0.75 [0.68, 0.81]} & \textbf{0.60 [0.30, 0.70]} \\
Mistral Small 3.2 & \textbf{0.50 [0.50, 0.87]} & \textbf{1.00 [0.50, 1.00]} & 0.41 [-0.14, 0.49] & 0.50 [-0.50, 0.50] & \textbf{0.30 [0.16, 0.42]} & \textbf{0.54 [0.19, 0.57]} \\
Gemma 3 27B & -0.50 [-1.00, 1.00] & 0.50 [-1.00, 1.00] & \textbf{0.67 [0.67, 0.67]} & 0.50 [-0.50, 1.00] & \textbf{0.39 [0.27, 0.50]} & 0.31 [-0.09, 0.71] \\
\midrule
Model average & \textbf{0.71 [0.57, 0.82]} & \textbf{0.67 [0.44, 0.80]} & \textbf{0.76 [0.71, 0.77]} & \textbf{0.57 [0.21, 0.64]} & \textbf{0.51 [0.48, 0.54]} & \textbf{0.65 [0.54, 0.67]} \\
\bottomrule\end{tabular}}\end{table*}

\subsection{Actionable Recommendation}
\label{sec:actionable_recommendation}

Actionable recommendation is dominated by research in algorithmic recourse and finance~\cite{karimi2022survey,keane2021if}, with extensions to healthcare~\cite{lacerda2023algorithmic}. 
Two aspects matter: (1) whether users follow suggested recourses, and (2) whether the recourse improves their situation if followed, with consideration to e.g. causality~\cite{karimi2021algorithmic}, robustness~\cite{dominguez2022adversarial}, and diversity~\cite{mothilal2020explaining}.
As our focus is human-AI interaction, we target the former. 
We replicate Tominaga et al.~\cite{tominaga2024reassessing}, who placed a Japanese cohort in a personalised loan-rejection scenario in which each participant rated two sparse and three random recourses. 
Their main finding was that $L_p$-norm cost moderately correlates with willingness to act (Figure~\ref{fig:results2}(A)). 
In our framework, $H$ are non-experts, $AI$ is a rule-based classifier, $\{\xi_i\}_{i=1}^N$ are recourses categorized by $L_p$ distance, and $\phi(\cdot)$ is a 1--7 Likert willingness rating; the ``Benefit'' term in Equation~\ref{eq:recourse} defaults to a constant since only propensity to act is measured.

We adapted the loan-rejection task using German Credit data because the original study materials were unavailable.
Due to inconsistencies in datasets, we omit the $L_0$ analysis from our main results, but include them in Figure~\ref{fig:actionable_rec_all_results1} for transparency.
We generated 200 distinct recourse plans for 39 rejected applicants, retaining only plans that changed rejection to acceptance under the same fitted random forest. 
Each MLLM was asked to rate its willingness to carry out each plan on a 1--7 scale. 
The main analysis correlates $L_1$ distance with willingness, using bootstrap resampling conditional on the observed applicant pool.

Table~\ref{tab:results_all} shows positive negative-$L_1$--willingness correlations for all 14 models, with point estimates ranging from $\rho=0.15$ to $0.76$ and all intervals entirely above zero. 
Thus, higher-cost plans generally receive lower willingness ratings. 
Figures~\ref{fig:actionable_rec_all_results1} and~\ref{fig:actionable_rec_all_results2} show fitted willingness curves, accounting for repeated plans from the same applicant.

\subsection{System Engagement}
\label{sec:system_eng}

Explanations have long been used to drive user engagement with AI, with the most prominent example being recommender systems~\cite{mcsherry2005explanation,gedikli2014should}. 
We focus on Herlocker et al.~\cite{herlocker2000explaining}, who tested 21 explanation types for convincing users to act on a MovieLens recommendation, asking participants how likely they were to see the film on a 1--7 Likert scale. 
Nearest-neighbour explanations performed best, feature-based explanations (actor, awards) slightly worse, and statistical explanations worst overall. 
In our framework, $H$ are non-experts, $AI$ is a collaborative filtering recommender, $\{\xi_i\}_{i=1}^{21}$ are the explanation types, and $\phi(\cdot)$ is the Likert rating.

We adapted six explanation types documented by Herlocker et al.~\cite{herlocker2000explaining}, including a no-explanation control, and substituted a fixed placeholder movie title. 
Each model generated 77 response blocks, with one 1--7 likelihood rating per condition in each block, yielding 462 responses. 
We retained complete six-condition blocks for analysis and compared their condition means with the published human means, preserving the human tie between two conditions. 

Engagement has the highest average correlation across models in our benchmark (Table~\ref{tab:results_all}). 
All 14 models have positive point estimates, ranging from $\rho=0.41$ to $0.90$, and 13 have intervals entirely above zero. 
GPT-5 mini and o4-mini share the highest point estimate ($\rho=0.90$, 95\% CI $[0.84,0.90]$).
Figure~\ref{fig:results2}(C) illustrates the relationship between human and MLLM mean ratings; the full model breakdown appears in Figure~\ref{fig:sys_eng_all_results}. 
These ratings measure stated viewing intention rather than observed engagement as these are shown to correlate~\cite{sheeran2002intention}.

\section{Sampling stability}

\begin{wrapfigure}{r}{0.49\textwidth}
    \centering
    \includegraphics[width=\linewidth]{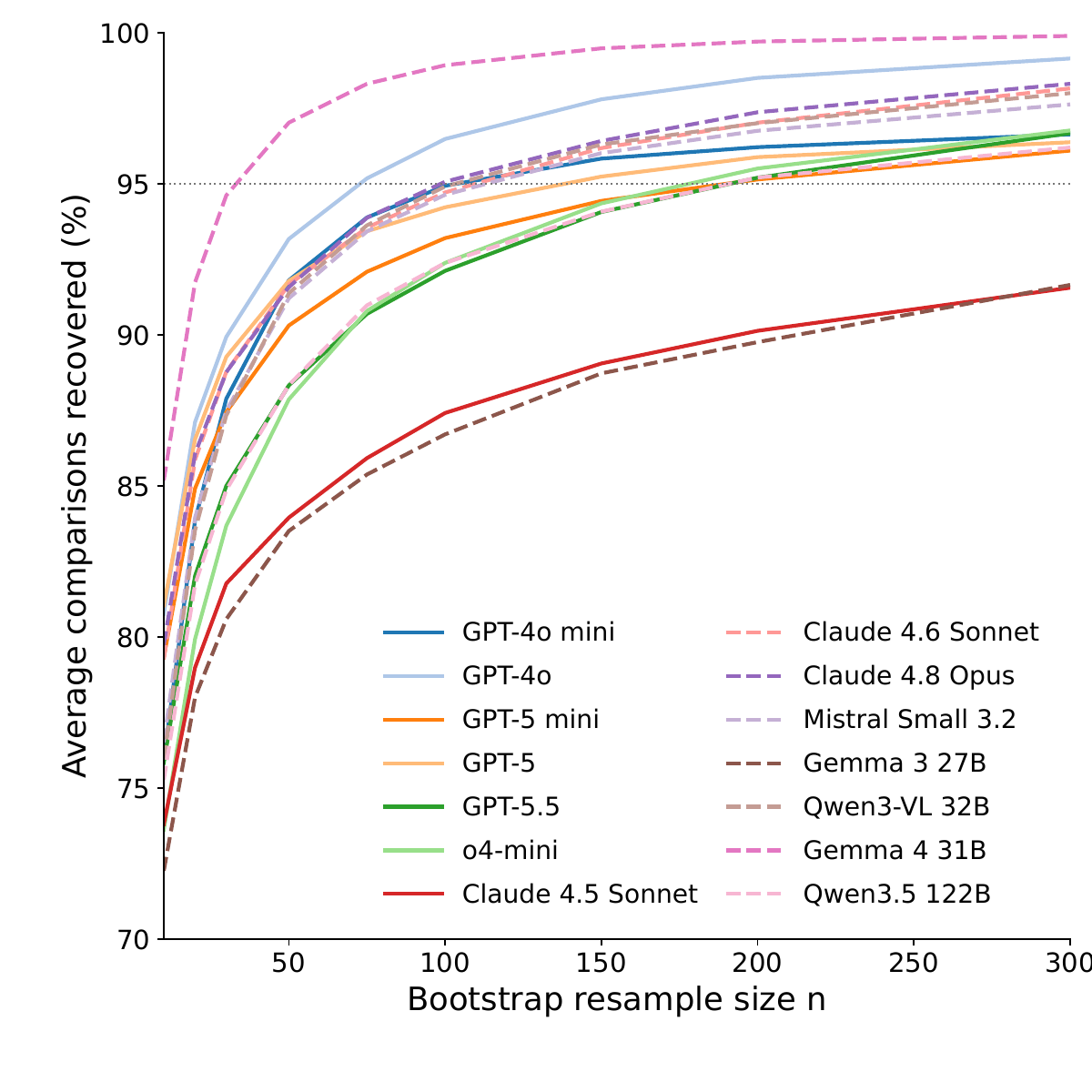}
    \caption{Pairwise ranking stability by model, averaged across
    five studies. 
    Curves show recovery of observed comparison
    directions under bootstrap resampling; the dotted line marks
    95\% recovery.}
    \label{fig:ranking_convergence}
\end{wrapfigure}

An important consideration for practitioners of our framework is what MLLM sample size is necessary to achieve reliable rankings for explanation methods.
Figure~\ref{fig:ranking_convergence} addresses this by examining how sample size affects the stability of the MLLM rankings observed in our main tests. 
For each model and study, we generated 10,000 bootstrap resamples using our existing data at sizes 1-300 and measured the fraction of pairwise condition orderings preserved relative to the full-data MLLM ranking. 
We averaged these scores across all studies, with recourse instead measuring preservation of the sign of the correlation between negative $L_1$ distance and willingness to act.

Average recovery across models increased from 93.4\% at $n=100$ to 96.7\% at $n=300$. 
These results suggest that 100 independent evaluation units provide a practical starting budget for recovering most comparison directions. 
However, some models such as Claude 4.5 Sonnet and Gemma 3 27B performed relatively poorly, and require many more samples for reliable rankings.
In contrast, the fastest converging model was Gemma 4 31B.

\section{Related Work}
There has been increasing investigation into whether LLMs can replicate human user study responses~\cite{dillion2023can,zhong2025synthetic,cui2024can}.
Cui et al.~\cite{cui2024can} showed that MLLMs can replicate the rankings of methods across large-scale experiments, but they were restricted to a single model (GPT-4) and psychological studies.
De Bona et al.~\cite{deevaluating} provided the first proof-of-concept that LLMs could replicate conclusions from an XAI user study, operationalising alignment as ANOVA effect replication in a single 2×2 study. 
Three aspects limit the generalisability of their findings however; they evaluate a single study and domain, their alignment metric (effect replication) does not directly address the question of ranking multiple XAI candidates, which is what practitioners need, and they use models that---as they themselves note---are not representative of current frontier capabilities. 
We extend this line of work by evaluating multiple candidate explanations across five purpose-grounded studies and 14 MLLMs, including nine API and five open-weight models.
We report rank correlations alongside bootstrap intervals and examine how agreement with human study findings varies by model and task.
Our conceptual analysis provides sufficient conditions for rank preservation in terms of condition-specific calibration error relative to the human mean difference (Section~\ref{sec:theory}), and we provide guidance on the sample size needed by varying model families in Figure~\ref{fig:ranking_convergence}.


\section{Discussion}
\label{sec:discussion}

Here we discuss potential caveats and a practical workflow for our MLLM human surrogate evaluation framework before concluding.
First, MLLMs are black-box evaluators, which limits interpretation of their responses. 
Our results and prior work~\cite{cui2024can,deevaluating} indicate that they can reproduce some comparative human-study outcomes, with agreement depending on the model and task. 
Returned reasoning text supports behavioral inspection of how a model describes the supplied explanations and feedback, as illustrated by our post-hoc audit on the model improvement study (see Appendix~\ref{app:mi-case-analysis}). 
However, it does not provide verified access to internal computation or establish that humans and models reach their conclusions through similar processes. 
Investigating those processes would require dedicated mental-model elicitation~\cite{hoffman2023measures} and mechanistic analyses~\cite{marks2025auditing}.
As an initial step, our recourse-rationale analysis (see Appendix~\ref{app:recourse-rationales}) provides preliminary evidence of mental model alignment between MLLM chain-of-thought reasoning and human written thought processes.

Another valid concern is that public datasets may appear in MLLM pre-training data, conflating results with memorization. 
However, eliciting such memorization is non-trivial~\cite{carlini2022quantifying}, and in our framework this is only a meaningful concern for reliance calibration and knowledge extraction, where predictions on test data are required. 
The latter can be addressed through novel datasets, as we demonstrated. 
For the former, as our framework only modifies the explanation and all other confounders are controlled between conditions, it is less of an issue. 
To elaborate, even if the MLLM's base accuracy is artificially inflated by memorization, this would (in theory) hold across all conditions, leaving the relative effect of the explanation intact.
However, this would only be true if sufficient headroom remains; evidence of this is seen in Figure~\ref{fig:reliance_all_results_bios} with the NLP reliance study. 
Appendix~\ref{app:memorization_tests} provides additional evidence that memorization was unlikely a significant confounder.

Lastly, our benchmark includes nine API models from two model families which may receive updates that cause our results to drift.
However these drifts are largely seen to occur in waves before returning to baseline~\cite{tschisgale2026daily}.
Nevertheless, to help address this, we ran our results with five open-weight models which will never drift, provide raw data and chain-of-thought responses from all our MLLMs, and a reproducible code base.
Broader model coverage remains valuable however.

A practical workflow for our framework would likely benefit from the following approach:

\begin{enumerate}
\setlength{\itemsep}{2pt}
\setlength{\parsep}{0pt}

\item \textbf{Define comparison.}
Identify which purpose-grounded area and associated metric your application best fits in Figure~\ref{fig:title}. 
Then, design prompts that do a controlled comparison between candidate explanations and a no-explanation control (either within or between subjects design).

\item \textbf{Check suitability.}
The current evidence suggests that structurally distinct explanation types which give the largest differences in human studies give the most reliable rankings in our framework (see Appendix~\ref{app:human_means_vs_rankings}).
As such, any measures to estimate the effect sizes in user testing would be warranted to help interpret the final results and plan evaluations.
In addition, text classification tasks seem vulnerable to a type of ceiling effect due to the ability of MLLMs to perform well at the task with no assistance and should be approached with additional caution.

\item \textbf{Model selection.}
The evidence from Table~\ref{tab:results_all} provides a starting point for model selection.
GPT models and larger open-weight Qwen variants are reasonable starting candidates, subject to computational budget. 
However, suitability varies by task, and parameter count alone
does not establish performance.
Where feasible, it is likely best to assess sensitivity across different model families.

\item \textbf{Sample size and uncertainty.}
Continue sampling until bootstrap confidence intervals resolve the
comparisons of interest, supporting either a meaningful directional
difference or practical equivalence within a prespecified tolerance.
Account for repeated checks and multiple comparisons when constructing
these intervals. Our convergence analysis suggests reasonable
starting budgets within our datasets (see Figure~\ref{fig:ranking_convergence}).
This provides a simple starting heuristic rather than a validated sample-size requirement for new studies.

\item \textbf{Use the evidence proportionately.}
Complement existing metrics with these comparisons and optionally
pilot shortlisted methods with humans based on the results. 
Averaging results over several prompt variations would be beneficial for robustness.

\end{enumerate}

To conclude, our results suggest that MLLM-based evaluation is a meaningful complement for XAI evaluation, as all models trended positively overall and most converge quickly (see Table~\ref{tab:results_all} and Figure~\ref{fig:ranking_convergence}). 
The ranking phenomenon is robust enough to lend practical credibility, yet fragile enough in specific conditions to warrant caution, as discussed in Section~\ref{sec:theory} and demonstrated during the model improvement and NLP reliance studies. 
We expect our results not to replace user testing, but instead to complement existing evaluation methodologies and help move the field away from relying heavily on proxy metrics which struggle to evaluate an explanation's true utility.

\section*{DISCLAIMER}
This paper was prepared for informational purposes by the
Artificial Intelligence Research group of JPMorgan Chase \&
Co., and its affiliates (“JP Morgan”) and is not a product of
the Research Department of JP Morgan. JP Morgan makes
no representation, and warranty whatsoever, and disclaims
all liability, for the completeness, accuracy or reliability
of the information contained herein. This document is not
intended as investment research or investment advice, or a
recommendation, offer or solicitation for the purchase or
sale of any security, financial instrument, financial product
or service, or to be used in any way for evaluating the merits
of participating in any transaction, and shall not constitute a
solicitation under any jurisdiction or to any person, if such
solicitation under such jurisdiction or to such person would
be unlawful.
© 2026 JP Morgan Chase \& Co. All rights reserved.

\bibliographystyle{plain}
\bibliography{biblo}

\newpage

\appendix
\onecolumn
\section{MLLM replication studies}
\label{app:replication_details}
%

\subsection{Models Used}
\label{app:model_ids}


\begin{center}
\small
\begin{tabular}{@{}p{0.27\linewidth}p{0.69\linewidth}@{}}
\toprule
Model & Snapshot, checkpoint or inference-profile identifier \\
\midrule
\multicolumn{2}{@{}l}{\textit{API models}} \\
GPT-4o & \path{gpt-4o-2024-11-20} \\
GPT-4o mini & \path{gpt-4o-mini-2024-07-18} \\
o4-mini & \path{o4-mini-2025-04-16} \\
GPT-5 & \path{gpt-5-2025-08-07} \\
GPT-5 mini & \path{gpt-5-mini-2025-08-07} \\
GPT-5.5 & \path{gpt-5.5-2026-04-24} \\
Claude 4.5 Sonnet &
\path{anthropic.claude-sonnet-4-5-20250929-v1:0} \\
Claude 4.6 Sonnet & \path{NA} \\
Claude 4.8 Opus & \path{NA} \\
\midrule
\multicolumn{2}{@{}l}{\textit{Open-weight models}} \\
Gemma 3 27B &
\path{pytorch/gemma-3-27b-it-FP8} \\
Gemma 4 31B &
\path{RedHatAI/gemma-4-31B-it-FP8-dynamic} \\
Qwen3-VL 32B &
\path{Qwen/Qwen3-VL-32B-Instruct-FP8} \\
Qwen3.5 122B &
\path{Qwen/Qwen3.5-122B-A10B-GPTQ-Int4} \\
Mistral Small 3.2 &
\path{stelterlab/Mistral-Small-3.2-24B-Instruct-2506-FP8} \\
\bottomrule
\end{tabular}
\end{center}

\subsection{Appropriate reliance}
We adapted the income-prediction and biography-classification tasks of Chen et al.~\cite{chen2023understanding} using Adult Census and BIOS data.
We reconstructed the selected cases (as the originals were not available), AI predictions and explanations from the available methodological information, including the target proportion of correctly classified cases.
These are adaptations of the study materials rather than exact reproductions of every original participant's task.
The published mean orderings provide our human references.

\subsubsection{Adult Dataset Experiment}

The Adult task uses nine displayed features: \texttt{age}, \texttt{workclass}, \texttt{education}, \texttt{marital-status}, \texttt{occupation}, \texttt{native-country}, \texttt{hours-per-week}, \texttt{gender}, and \texttt{race}.
The target is whether income exceeds \$50,000; the source uses \texttt{income} as the target column.
Categorical predictors are encoded for model fitting and displayed as readable category names in prompts.

The reconstructed classifier is a Random Forest with 100 estimators, maximum depth 10, and the dataset-block seed as its random state.
Adult uses an 80/20 stratified train/test split.
The case-selection procedure targets an experienced AI accuracy of 62.5\%, rather than the classifier's accuracy over its entire test bank.
This was done to reflect the original study.
Nearest-neighbour candidate search considers 50 training cases using Euclidean distance.
Feature-based explanations use LIME and refer to the displayed AI prediction.
The frozen materials correct the income-label mapping and present all five labelled demonstrations.

\subsubsection{BIOS Dataset Experiment}

The BIOS task uses biography text from \texttt{hard\_text}, restricted to five professions: physician, professor, psychologist, surgeon, and teacher.
We map the source profession codes to these labels and use the supplied training and test partitions.
Text is represented by a \texttt{CountVectorizer} with English stop words and at most 5,000 features.
The classifier is a Random Forest with 100 estimators, maximum depth 10, and the dataset-block seed as its random state.
Nearest-neighbour candidate search uses Euclidean distance over the dense bag-of-words vectors and considers 50 training cases.
LIME text explanations describe the displayed AI prediction.

The reconstruction targets 62.5\% AI accuracy on the selected cases and approximately 63\% agreement between the first neighbour's true label and the test case's true label.
These are selection targets, not guaranteed rates in every exported cohort.
The realised AI accuracy is 260/416 for GPT-4o, GPT-4o mini, GPT-5 mini and o4-mini, and 263/416 for the remaining models.
The prompt labels and predictions use the same five-profession mapping.

\subsubsection{Experiment methodology}
Each model is evaluated on 26 dataset-seed blocks per domain, with 16 cases in each block.
Every case is presented under all three conditions, yielding 416 responses per condition and domain, with one response per case--condition pair.
Each prompt includes five fixed labelled demonstrations; these are shared across conditions rather than resampled for every query.

\begin{enumerate}
    \item \textbf{Control:} the case and demonstrations, with no AI prediction or explanation.
    \item \textbf{Saliency:} the same case and demonstrations, the AI prediction, and a LIME explanation.
    \item \textbf{Nearest neighbours:} the same case and demonstrations, the AI prediction, and two training cases with their true labels and AI predictions.
\end{enumerate}
Consequently, comparisons with control measure the combined contribution of AI advice and explanation; they do not isolate explanation utility from advice alone.
Conditions are matched within each model.

Accuracy is the proportion of responses with the correct class label; invalid answers count as incorrect.
The primary table computes Spearman's $\rho$ from the full-data condition means and obtains percentile intervals from 10,000 matched block-bootstrap resamples at the observed 26-block sample size.
Adult and BIOS are averaged within each reliance resample.
The human reference orderings are nearest neighbours $>$ control $>$ saliency for Adult, and nearest neighbours $>$ saliency $>$ control for BIOS.

\begin{table}[!ht]
\centering
\caption{Control prompt for appropriate reliance}
\begin{tabular}{@{}l@{}}
\toprule
\multicolumn{1}{c}{\textbf{Control prompt for appropriate reliance}}                                                                                                                                                                                                                                                                                                   \\ \midrule
\begin{tabular}[c]{@{}p{0.94\linewidth}@{}}

        Pretend you are a participant in a user study with some \\technical machine learning knowledge being asked to assist in a\\
        study to examine people's reliance on AI systems with or \\without explanations for the AI's predictions.\\
\\
        The domain is Adult Census and the features in order are: \texttt{\{feature\_names\}} \\
\\
        The possible labels are: \texttt{\{class\_names\}} \\
\\
        Here are 5 randomly sampled examples of training data to help \\you understand the domain: \texttt{\{five\_demonstration\_instances\}},\\
        and here are their labels: \texttt{\{five\_demonstration\_labels\}}\\
\\
        Your task is to look at a new test instance, and decide what you think the real label is.\\
\\
        Here is the new test instance: \texttt{\{row.test\_instance\}}\\
\\
        Outline your reasoning process step by step, and give your \\final answer as one of the possible class labels in the tags \\\texttt{<answer>...</answer>}.\\

\end{tabular} \\ \bottomrule
\end{tabular}
\label{tab:reliance_control}
\end{table}

\begin{table}[!ht]
\centering
\caption{Saliency prompt for appropriate reliance}
\begin{tabular}{@{}l@{}}
\toprule
\multicolumn{1}{c}{\textbf{Saliency prompt for appropriate reliance}}                                                                                                                                                                                                                                                                                                   \\ \midrule
\begin{tabular}[c]{@{}p{0.94\linewidth}@{}}

        Pretend you are a participant in a user study with some \\technical machine learning knowledge being asked to assist in a\\
        study to examine people's reliance on AI systems with or \\without explanations for the AI's predictions.\\
\\
        The domain is Adult Census and the features in order are: \texttt{\{feature\_names\}} \\
\\
        The possible labels are: \texttt{\{class\_names\}} \\
\\
        Here are 5 randomly sampled examples of training data to help \\you understand the domain: \texttt{\{five\_demonstration\_instances\}},\\
        and here are their labels: \texttt{\{five\_demonstration\_labels\}}\\
\\
        Your task is to look at a new test instance, and decide what you think the real label is.\\
\\
        Here is the new test instance: \texttt{\{row.test\_instance\}}\\
\\
        Here is the AI prediction for the \\new test instance: \texttt{\{row.test\_predicted\_label\}} \\
\\
        In addition, here is a feature saliency explanation for the AI prediction\\ where you see each feature, \\
        and its contribution (positive or negative) towards the \\prediction: \texttt{\{round\_numbers\_in\_string(row.lime\_explanation)\}}   \\ 
\\
        Outline your reasoning process step by step, and give your \\final answer as one of the possible class labels in the tags \\\texttt{<answer>...</answer>}.\\

\end{tabular} \\ \bottomrule
\end{tabular}
\label{tab:reliance_saliency}
\end{table}

\begin{table}[!ht]
\centering
\caption{Example-based prompt for appropriate reliance}
\begin{tabular}{@{}l@{}}
\toprule
\multicolumn{1}{c}{\textbf{Example-based prompt for appropriate reliance}}                                                                                                                                                                                                                                                                                                   \\ \midrule
\begin{tabular}[c]{@{}p{0.94\linewidth}@{}}

        Pretend you are a participant in a user study with some \\technical machine learning knowledge being asked to assist in a\\
        study to examine people's reliance on AI systems with or \\without explanations for the AI's predictions.\\
\\
        The domain is Adult Census and the features in order are: \texttt{\{feature\_names\}} \\
\\
        The possible labels are: \texttt{\{class\_names\}} \\
\\
        Here are 5 randomly sampled examples of training data to help \\you understand the domain: \texttt{\{five\_demonstration\_instances\}},\\
        and here are their labels: \texttt{\{five\_demonstration\_labels\}}\\
\\
        Your task is to look at a new test instance, and decide what you think the real label is.\\
\\
        Here is the new test instance: \texttt{\{row.test\_instance\}}\\
\\
        Here is the AI prediction for the \\new test instance: \texttt{\{row.test\_predicted\_label\}}
\\
        In addition, here is an explanation for the prediction in the \\form of the two nearest neighbors from the training data\\
        that the AI learned from: \\
\\
        The first nearest neighbor is \texttt{\{row.neighbor1\_instance\}},\\
        its label is '\texttt{\{row.neighbor1\_true\_label\}}',\\
        and the AI's prediction of this nearest neighbor is '\texttt{\{row.neighbor1\_predicted\_label\}}'\\
        The second nearest neighbor is \texttt{\{row.neighbor2\_instance\}},\\
        its label is '\texttt{\{row.neighbor2\_true\_label\}}',\\
        and the AI's prediction of this nearest neighbor is '\texttt{\{row.neighbor2\_predicted\_label\}}'\\
\\
        Outline your reasoning process step by step, and give your \\final answer as one of the possible class labels in the tags \\\texttt{<answer>...</answer>}.\\

\end{tabular} \\ \bottomrule
\end{tabular}
\label{tab:reliance_nns}
\end{table}

The human reference orderings are based on reported condition means and do not imply that every pairwise difference was statistically significant.
The descriptive accuracy plots do not establish replication of non-significant human comparisons.

\subsection{Human-machine teaching}
\label{app:hmt_setup}
Warren et al.~\cite{warren2024categorical} studied whether counterfactual and causal explanations help people learn a domain with a known outcome rule.
We use a synthetic adaptation rather than the Widmark-based domain, whose widely documented formula may be familiar to an MLLM.
Our domain has five anonymised features: two binary variables and three positive continuous variables.

\begin{figure}[!ht]
\centering
  \includegraphics[width=0.5\textwidth]{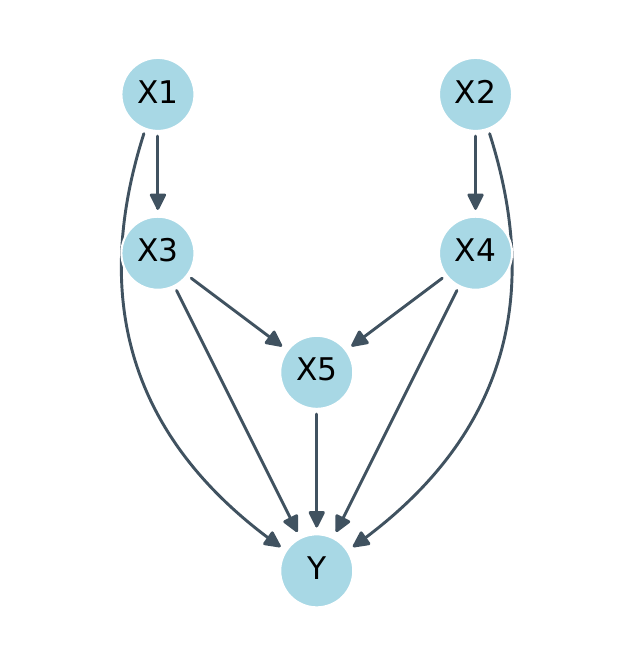}
  \caption{
  The five-feature causal graph used in our synthetic adaptation of Warren et al.~\cite{warren2024categorical}. All five features contribute to the outcome $Y$.
  }
  \label{fig:causal_graph}
\end{figure}

Let $X_1,X_2\sim\operatorname{Bernoulli}(0.5)$ independently, and let $Z_3,Z_4,Z_5$ be independent standard normal variables.
The continuous features are generated sequentially as
\begin{align*}
X_3 &= \operatorname{round}_2\!\left[\exp(0.5+0.8X_1+0.3Z_3)\right],\\
X_4 &= \operatorname{round}_2\!\left[\exp(0.7+X_2+0.3Z_4)\right],\\
X_5 &= \operatorname{round}_2\!\left[\exp\{0.1\ln(X_3+\epsilon)+0.1\ln(X_4+\epsilon)+0.3Z_5\}\right],
\end{align*}
where $\operatorname{round}_2$ rounds to two decimal places and $\epsilon=10^{-9}$ stabilises the logarithms.
The outcome is $Y=1$ when $S\geq4.71$, and $Y=0$ otherwise, where
\begin{align*}
S={}&X_1+X_2+1.2\ln(X_3+\epsilon)+\ln(X_4+\epsilon)+1.5X_5\\
&+0.1X_1\ln(X_3+\epsilon)+0.1X_2\ln(X_4+\epsilon).
\end{align*}
The graph contains $X_1\rightarrow X_3$, $X_2\rightarrow X_4$, $X_3\rightarrow X_5$, and $X_4\rightarrow X_5$, with all five features contributing to $Y$ (Figure~\ref{fig:causal_graph}).
The graph and equations define the task but are not supplied directly in the teaching prompts.

The five demonstrations are fixed and shared across models and conditions; their counterfactuals are fixed and supplied only in the counterfactual condition.
They include three negative and two positive examples, with one selected intervention for each feature.
The negative example $(1,0,2.63,2.69,0.80)$ breaks the shortcut $Y=X_1$.
The frozen original/counterfactual pairs, in feature order $(X_1,\ldots,X_5)$, are
\begin{center}\small
\begin{tabular}{@{}clclc@{}}
\toprule
Intervention & Original & $Y$ & Counterfactual & $Y'$\\
\midrule
$X_1$ & $(0,1,1.43,5.79,0.74)$ & 0 & $(1,1,3.18,5.79,0.80)$ & 1\\
$X_2$ & $(0,0,2.25,1.71,1.09)$ & 0 & $(0,1,2.25,4.65,1.20)$ & 1\\
$X_3$ & $(1,0,2.63,2.69,0.80)$ & 0 & $(1,0,3.17,2.69,0.82)$ & 1\\
$X_4$ & $(1,1,4.06,7.86,1.52)$ & 1 & $(1,1,4.06,0.46,1.14)$ & 0\\
$X_5$ & $(1,0,3.88,2.00,1.37)$ & 1 & $(1,0,3.88,2.00,0.83)$ & 0\\
\bottomrule
\end{tabular}
\end{center}
Counterfactual construction intervenes on one selected feature, propagates the change to its descendants while holding the structural noise terms fixed, and recomputes $Y$.
Binary interventions flip the selected value; continuous interventions search the 0.01 grid for the nearest selected-feature value that flips the label.
Thus a single intervention can change several displayed features.
The released runner uses the frozen pairs rather than regenerating them for each query.

The control receives the five original labelled instances.
The counterfactual condition additionally receives the paired label-flipping instances.
The causal condition instead adds the sentence:
``In this specific example, part of the reason the label is [label] is because feature [feature] was [value].''
The named feature is the corresponding intervention feature; no numerical feature-importance scores are provided.

The runner generates 10,000 source instances with seed 42 and an 80/20 stratified split.
It generates a fresh pool of 20,000 instances with seed 20260912, removes feature-identical matches to the training set and demonstration counterfactuals, and samples 300 questions without replacement using seed 42.
This is simple random test sampling with natural class balance, not a balanced test set.
Each question is presented in all three conditions with one response per condition, for 900 responses per model.
Every prompt ends with:
``Explain your reasoning briefly, and give your final answer as 0 or 1 in the tags \texttt{<answer>...</answer>}.''
The word `briefly', was chosen to avoid models giving overly verbose responses, especially for the counterfactual condition which frequently reached token limits due to the complexity.
Accuracy includes invalid answers as incorrect.
The primary table uses full-data condition means and 10,000 bootstrap resamples of the 300 matched questions, with the fixed human ordering counterfactual $>$ causal $>$ control.

\begin{table}[!ht]
\centering\small
\caption{Control prompt for human-machine teaching.}
\label{tab:teaching_control}
\begin{tabular}{@{}p{0.95\linewidth}@{}}
\toprule
Pretend you are a participant in a user study designed to (1) teach you about a complex causal domain, and (2) get you to make a prediction on a novel test instance given what you learned.\\
The data domain has 5 features and is structured as follows: - X1, X2: Binary variables (0 or 1) - X3, X4, X5: Continuous POSITIVE variables (representing quantities > 0) - Y: Outcome (0 or 1)\\
You are now going to learn about the underlying domain by studying the following information: Here are [len(examples)] training examples, and their labels Y to help you understand the domain: [formatted\_examples]\\
Here is the new test instance: [test\_instance]\\
Now, given all this information, make an informed prediction on what Y is for this test instance. Explain your reasoning briefly, and give your final answer as 0 or 1 in the tags <answer>...</answer>.\\
\bottomrule
\end{tabular}
\end{table}

\begin{table}[!ht]
\centering\small
\caption{Causal prompt for human-machine teaching.}
\label{tab:teaching_causal}
\begin{tabular}{@{}p{0.95\linewidth}@{}}
\toprule
Pretend you are a participant in a user study designed to (1) teach you about a complex causal domain, and (2) get you to make a prediction on a novel test instance given what you learned.\\
The data domain has 5 features and is structured as follows: - X1, X2: Binary variables (0 or 1) - X3, X4, X5: Continuous POSITIVE variables (representing quantities > 0) - Y: Outcome (0 or 1)\\
You are now going to learn about the underlying causal graph by studying the following explanation: Here are [len(examples)] training examples, and a causal explanation for their classification, and their labels Y to help you learn the underlying causal graph, study this carefully: [formatted\_examples]\\
Here is the new test instance: [test\_instance]\\
Now, given all this information, and what you have learned from the explanation, make an informed prediction on what Y is for this test instance. Explain your reasoning briefly, and give your final answer as 0 or 1 in the tags <answer>...</answer>.\\
\bottomrule
\end{tabular}
\end{table}

\begin{table}[!ht]
\centering\small
\caption{Counterfactual prompt for human-machine teaching.}
\label{tab:teaching_counterfactual}
\begin{tabular}{@{}p{0.95\linewidth}@{}}
\toprule
Pretend you are a participant in a user study designed to (1) teach you about a complex causal domain, and (2) get you to make a prediction on a novel test instance given what you learned.\\
The data domain has 5 features and is structured as follows: - X1, X2: Binary variables (0 or 1) - X3, X4, X5: Continuous POSITIVE variables (representing quantities > 0) - Y: Outcome (0 or 1)\\
You are now going to learn about the underlying causal graph by studying the following explanation: Here are [len(examples)] training examples, counterfactual versions of them where Y changes to the opposite class, and their labels Y to help you learn the underlying causal graph, study this carefully: [formatted\_examples]\\
Here is the new test instance: [test\_instance]\\
Now, given all this information, and what you have learned from the explanation, make an informed prediction on what Y is for this test instance. Explain your reasoning briefly, and give your final answer as 0 or 1 in the tags <answer>...</answer>.\\
\bottomrule
\end{tabular}
\end{table}

\subsection{System engagement}
We adapted six recommendation explanations from Herlocker et al.~\cite{herlocker2000explaining}: explanation IDs 1, 2, 4, 5, 11 and 21.
We chose these six because the original materials have been lost and the authors could not supply us with them, so we used all explanations which were detailed in the published paper only.
The prompt presents a user with \$7 and a free evening and asks how likely they are to see a recommended movie on a 1--7 scale.
The movie title is replaced by a fixed placeholder to reduce title-specific preferences; the recommendation explanations are supplied as text.
Explanation 11 is the no-explanation control.
Explanation 4 is described as a histogram in our prompts, whereas the original study presented a table of neighbour ratings; this is a presentation adaptation.

We collected 77 response blocks per model, with one response to each of the six conditions in each block: 462 responses per model.
These are repeated model samples, not 77 independent human participants.
We parse an explicit 1--7 answer and retain complete six-condition blocks for the comparative analysis.
The published human reference means, in explanation-ID order, are 5.25, 5.19, 4.97, 4.97, 4.53 and 3.94, preserving the tie between explanations 4 and 5.
The descriptive plots report Pearson's $r$ between the six human and model condition means; the primary table instead reports full-data Spearman's $\rho$ with 10,000 matched response-block bootstrap resamples at the observed complete-block count.
The six condition means are the correlation observations, not the number of response blocks.

\begin{table}[!ht]
\centering
\caption{Prompt for System Engagement}
\begin{tabular}{@{}l@{}}
\toprule
\multicolumn{1}{c}{\textbf{Prompt for System Engagement}}                                                                                                                                                                                                                                                                                                   \\ \midrule
\begin{tabular}[c]{@{}p{0.94\linewidth}@{}}

        Pretend you are a human user of a movie recommender system.\\
\\
        Imagine that you have \$7 and a free evening\\
        coming up. You are considering going to the\\
        theater to see a movie, but only if there is a movie\\
        worth seeing. To determine if there is a movie\\
        worth seeing, you consult MovieLens for a\\
        personalized movie recommendation. MovieLens\\
        recommends one movie, and provides an explanation\\
        to justify the recommendation.\\
\\
        The movie recommended was: \texttt{\{movie\_title\}} \\
\\
        Here is its explanation for the recommendation: \texttt{\{explanation\}} \\
\\
        The movies are rated on a scale of 1-5.\\
        Based on the provided explanation, please indicate on a scale of 1-7\\
        how likely you are to see the movie. \\
\\
        First, outline your reasoning step by step, and put your answer\\ in between the tags \texttt{<answer>...</answer>}\\
\\
        It is very important you follow this exact format.\\

\end{tabular} \\ \bottomrule
\end{tabular}
\label{tab:system_engagement_prompt}
\end{table}

\begin{table}[!ht]
\centering
\caption{Prompt for System Engagement -- No Explanation}
\begin{tabular}{@{}l@{}}
\toprule
\multicolumn{1}{c}{\textbf{Prompt for System Engagement -- No Explanation}}                                                                                                                                                                                                                                                                                                   \\ \midrule
\begin{tabular}[c]{@{}p{0.94\linewidth}@{}}

        Pretend you are a human user of a movie recommender system.\\
\\
        Imagine that you have \$7 and a free evening\\
        coming up. You are considering going to the\\
        theater to see a movie, but only if there is a movie\\
        worth seeing. To determine if there is a movie\\
        worth seeing, you consult MovieLens for a\\
        personalized movie recommendation. MovieLens\\
        recommends one movie.\\
\\
        The movie recommended was: \texttt{\{movie\_title\}} \\
\\
        Please indicate on a scale of 1-7\\
        how likely you are to see the movie. \\
\\
        First, outline your reasoning step by step, and put your answer\\ in between the tags \texttt{<answer>...</answer>}\\
\\
        It is very important you follow this exact format.\\

\end{tabular} \\ \bottomrule
\end{tabular}
\label{tab:system_engagement_prompt_control}
\end{table}

\subsection{Actionable Recourse Experimental Setup}
\label{app:recourse_setup}

This experiment asks MLLMs to assess recourse plans generated from the German Credit dataset.
We preserve both requested ratings but analyse only \texttt{actionability\_score}, the stated willingness to carry out a plan.

\paragraph{Feature constraints and action rules.}
The mutable numeric features are \texttt{duration}, \texttt{credit\_amount} and \texttt{existing\_credits}; permitted edits reduce their integer values within training-derived bounds.
The mutable categorical features are \texttt{status} and \texttt{savings}, with adjacent codebook transitions:
\begin{itemize}
    \item \texttt{status}: A11 $\rightarrow$ A12, or A12 $\rightarrow$ A13.
    \item \texttt{savings}: A61 $\rightarrow$ A62, A62 $\rightarrow$ A63, or A63 $\rightarrow$ A64.
\end{itemize}
All other fields are held fixed, including credit history, employment tenure, housing, age and personal status.
Categories are rendered using the historical Statlog codebook rather than treated as ordinal numbers.
These constraints define the allowed plans; they do not establish that every plan is feasible for every applicant.

For an original profile $x$ and revised profile $x'$, distances are
\begin{align*}
L_0(x,x') &= \sum_j \mathbf{1}[x_j\ne x'_j],\\
L_1(x,x') &= \sum_{j\in\mathcal{N}}\frac{|x'_j-x_j|}{\sigma_j}
           +\sum_{j\in\mathcal{C}}\mathbf{1}[x_j\ne x'_j],
\end{align*}
where $\mathcal{N}$ and $\mathcal{C}$ denote numeric and categorical fields, and $\sigma_j$ is the numeric feature's training-set standard deviation.
$L_0$ counts changes to original fields rather than one-hot columns.

\paragraph{Action generation and selection.}
The generator uses a 70/30 stratified train/test split with seed 42, one-hot categorical encoding, and a Random Forest with 100 estimators, maximum depth 12 and random state 42.
It searches rejected test profiles over permitted categorical transitions and numeric reductions at predefined distance scales, retaining candidate plans with $L_1\leq13$.
Every retained plan must change rejection to acceptance under the same fitted classifier; no second classifier is used to enlarge the candidate pool.
We select 200 distinct plans using round-robin allocation across $L_0$ strata and occupied $L_1$ bins, with finer bins near zero.
The selected plans share 39 applicants and therefore are not 200 independent people.
The rendered prompt states the actual loan purpose, duration and amount, using Deutsche Mark (DM) units from the source dataset.

\paragraph{Aggregate analysis.}
We plot willingness separately against $L_0$ and $L_1$ using Gaussian generalized additive mixed models fitted by REML in R's \texttt{mgcv} package.
The model includes a thin-plate smooth of distance, an applicant random intercept, and an applicant random slope:
\[
\texttt{y \textasciitilde\ s(x,bs="tp",k=k) + s(uid,bs="re") + s(x,uid,bs="re")},
\]
where $k=\min(10,\text{number of distinct distance values})$.
Only valid integer ratings from 1 to 7 enter a fit.
Population curves exclude the applicant-specific effects; the bands are approximate pointwise 95\% intervals for the population mean, calculated as fitted mean $\pm1.96$ standard errors with smoothing-parameter uncertainty included.
They are not prediction intervals, and the fitted curves are not constrained to the rating range.

\begin{figure}[!ht]
\centering
  \includegraphics[width=\textwidth]{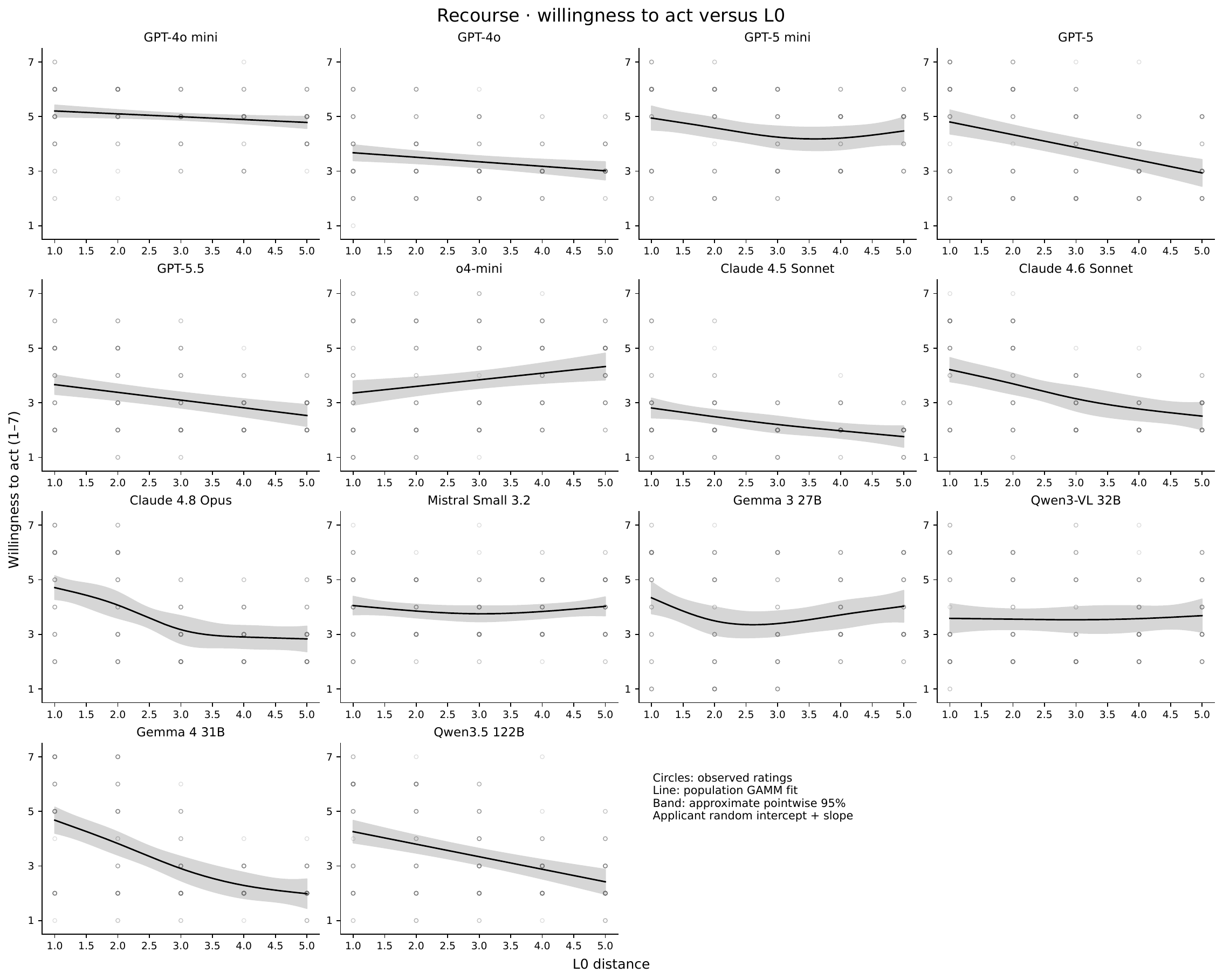}
        \caption{Willingness to act versus $L_0$ changed-feature count, retained as a supplementary diagnostic after its exclusion from the primary aggregate. 
        Points show valid ratings; curves are fitted population-mean GAMM trends, with approximate pointwise 95\% mean intervals. The 200 plans share 39 applicants. Our adaptation permits numerical and categorical changes, whereas Tominaga et al.~\cite{tominaga2024reassessing} represented many quantities using ordered categories. $L_0$ counts changed features without measuring the magnitude of each change: reducing a loan amount by 1~DM or 1000~DM contributes equally. Differences in feature representation and permitted changes limit direct comparison with the original study and motivated reporting $L_0$ separately. We retain these results for transparency.}  \label{fig:actionable_rec_all_results1}
        \end{figure}

The main table reports Spearman's correlation between negative $L_1$ and willingness, with plan-level bootstrap resampling conditional on the observed applicant pool.
This is a directional cost--willingness relationship, not direct agreement with matched human recourse ratings.
Excluding $L_0$ from the primary aggregate was a post-hoc decision: a small and a large numeric change count equally under $L_0$, creating a mismatch with the original study's feature representation.
The $L_0$ results are retained in Figure~\ref{fig:actionable_rec_all_results1}.
The visual-comparison prompt below is retained as a historical inspection procedure and does not determine the primary correlation results.

\begin{table}[!ht]
\centering\small
\caption{Prompt for rating recourse plans.}
\label{tab:recourse_prompt}
\begin{tabular}{@{}p{0.95\linewidth}@{}}
\toprule
Imagine you are a human applying for a loan. You are at a financial institution undergoing a loan assessment. An AI system determines approval or denial based on your profile data.\\
The requested loan's purpose, amount and duration are shown below. Monetary amounts are in Deutsche Mark (DM), as recorded in this historical dataset.\\
\textbf{YOUR CURRENT PROFILE:} \texttt{[profile with readable categories and units]}\\
An AI system has rejected your loan application. The following changes would make that same AI system accept your revised profile. All other profile details stay the same.\\
\textbf{AI Explanation:} \texttt{[numbered changes from original to proposed values]}\\
\textbf{Your Task:} Please answer the following two questions using an integer between 1--7 where 1=``Strongly no'', 7=``Strongly yes'', and 4=``Neutral''.\\
1. Is the AI System's plan a reasonable explanation for the rejection of your loan application?\\
2. Would you carry out the plan presented by the AI system to obtain loan approval?\\
Outline your reasoning and then give your scores in JSON like:
\texttt{\{"reasoning": string, "acceptance\_score": integer, "actionability\_score": integer\}}\\
Do not put anything except a single integer for the two scores.\\
\bottomrule
\end{tabular}
\end{table}


\subsection{Model Improvement}
\label{app:diabetes_optimization}
We evaluate whether MLLMs can improve a diabetes classifier by proposing training-data configurations under data-centric (DCE), model-centric (MCE) and hybrid (HYB) conditions.
Each run has five proposal attempts.
The final release contains 100 seed blocks per model and condition, except Qwen3.5 122B, which has 40 due to computational constraints.

\paragraph{Dataset and baseline.}
The released Pima Indians Diabetes CSV has 768 rows, target \texttt{Outcome}, and eight numeric features: \texttt{Pregnancies}, \texttt{Glucose}, \texttt{BloodPressure}, \texttt{SkinThickness}, \texttt{Insulin}, \texttt{BMI}, \texttt{DiabetesPedigreeFunction}, and \texttt{Age}.
The released open-weight routine uses an unstratified 85/15 split with the run seed, producing 652 training rows and 116 feedback rows.
The Random Forest uses 30 estimators, maximum depth 5, minimum split size 50, minimum leaf size 25, \texttt{max\_features=sqrt}, maximum 10 leaf nodes, and \texttt{random\_state=42}.
The split changes across seeds; the classifier random state remains fixed.
Conditions share the original partition within each seed.

\subsubsection{Optimization conditions: DCE, MCE, HYB}
Each proposal specifies whether to include each feature and optional minimum/maximum values for filtering training rows.
The configuration is applied afresh to the original training partition, not cumulatively to the previously filtered data.
The classifier is retrained and evaluated on the same fixed feedback partition.

DCE supplies class balance, up to three highly correlated feature pairs ($|r|\geq0.8$), per-feature summaries, and data-quality scores for zeros, outliers, balance and correlation.
The summaries include minimum, mean, maximum, fifth and 95th percentiles, zero frequency, IQR-based outliers, and a mean--median skewness classification.
These summaries and observed feature ranges use the full source CSV and remain static across attempts.
DCE and HYB receive the original static feature-distribution image; MCE does not receive it in the released open-weight routine.

MCE supplies the current best Random Forest's impurity-based feature importance and rules from a surrogate Decision Tree trained on its predictions over the original training rows for the current features.
The surrogate has maximum depth 4, minimum leaf size 20 and random state 42.
Up to four leaf rules per class are shown, sorted by supporting sample count and confidence.
HYB combines the DCE and MCE information.

\subsubsection{Prompt construction}
Each prompt contains the baseline and current best accuracy, their difference, the available features, the best configuration when present, condition-specific signals, observed feature ranges, and the previous attempts' configurations, scores and feedback.
The source prompt's displayed sample counts remain based on the full CSV rather than tracking filtered training-row counts.
The five-attempt interaction is our adaptation of iterative user-driven model improvement.
The prompt requests a reasoning section followed by a configuration JSON object (Table~\ref{tab:diabetes_prompt}).

The primary endpoint takes the highest successfully evaluated accuracy among attempts 1--5, excluding baseline from the maximum.
For baseline accuracy $a_0$ and best proposal accuracy $a_\star$, the reported relative improvement is
\[
100\,(a_\star-a_0)/a_0.
\]
This is a percentage change, not the percentage-point difference $100(a_\star-a_0)$; negative gains remain possible.
The source history's \texttt{final\_accuracy} instead stores the best-so-far score including baseline and is not used directly as this endpoint.
Repeated feedback and best-proposal selection use the same feedback partition, so the endpoint is not an untouched generalisation estimate.
For the primary table, full-data condition means are compared with the figure-based human ordering HYB $>$ DCE $>$ MCE, using 10,000 matched seed-bootstrap resamples at the observed sample size.

\begin{table}[!ht]
\centering
\caption{Optimization prompt for diabetes prediction (DCE/MCE/HYB)}
\begin{tabular}{@{}l@{}}
\toprule
\multicolumn{1}{c}{\textbf{Optimization prompt structure}} \\ \midrule
\begin{tabular}[c]{@{}p{0.94\linewidth}@{}}
\textbf{ROLE:} You are a medical professional optimizing a diabetes prediction ML model by configuring training data.\\
\\
\textbf{CURRENT STATUS}\\
- Baseline accuracy: \texttt{\{baseline\_accuracy\}}\\
- Current best accuracy: \texttt{\{current\_best\}}\\
- Improvement so far: \texttt{\{delta\}}\\
- Original samples: \texttt{\{n\_orig\}}, Current: \texttt{\{n\_current\}}\\
- Available features: \texttt{\{feature\_list\}}\\
\\
\textbf{CURRENT BEST CONFIGURATION}\\
\texttt{\{formatted\_config\}}\\
\\
\textbf{DATA-CENTRIC SIGNALS} (for DCE/HYB)\\
- Class balance: pos \%, neg \%\\
- Highly correlated pairs (if any)\\
- Data quality scores: zeros, outliers, balance, correlation, overall\\
- Per-feature summary: min, mean, max, p5, p95, skew, zeros \%, outliers \%\\
\\
\textbf{MODEL-CENTRIC SIGNALS} (for MCE/HYB)\\
- Feature importance ranking\\
- Top decision rules for Class 1 and 0 (confidence, samples)\\
\\
\textbf{OBSERVED FEATURE RANGES}\\
- \texttt{feat}: [min, max], ...\\
\\
\textbf{HISTORY}\\
\texttt{\{formatted\_recent\_iterations\}}\\
\\
\textbf{TASK}\\
Propose a NEW configuration to beat \texttt{\{current\_best\}}.\\
Return your \textbf{REASONING} followed by \textbf{CONFIGURATION} JSON:\\
\{ "FeatureName": \{ "include": true/false, "min": number\_or\_null, "max": number\_or\_null \} \}\\
\end{tabular} \\ \bottomrule
\end{tabular}
\label{tab:diabetes_prompt}
\end{table}

\section{Memorization tests}
\label{app:memorization_tests}
We conducted targeted elicitation checks to examine whether GPT-4o could reproduce published study results from memory.
These checks used the same \path{gpt-4o-2024-11-20} snapshot as the benchmark.
Failure to elicit a result does not establish that it was absent from training data or could not influence responses under another prompt.

\subsection{\texorpdfstring{System engagement: Herlocker  et al.~\cite{herlocker2000explaining}}{System engagement: Herlocker et al.}}
We asked GPT-4o to report the published mean rating for each tested explanation method from Herlocker et al.~\cite{herlocker2000explaining}, using Table~\ref{tab:memorization}.

\begin{table}[tbp]
\centering
\caption{The prompt used to try elicit memorization for our replication study of Herlocker  et al.~\cite{herlocker2000explaining}.}
\begin{tabular}{@{}l@{}}
\toprule
\multicolumn{1}{c}{\textbf{MLLM Prompt}}                                                                                                                                                                                                                                                                                                   \\ \midrule
\begin{tabular}[c]{@{}p{0.94\linewidth}@{}}In Herlocker et al.’s 2000 paper \\ ``Explaining Collaborative Filtering Recommendations'', \\ what did users rate \{\texttt{insert\_method\_name}\} on a 1–7 scale?\\ Give your answer in between the tags <answer>...</answer>
\end{tabular} \\ \bottomrule
\end{tabular}
\label{tab:memorization}
\end{table}

\begin{figure}[!ht]
\centering
  \includegraphics[width=0.5\textwidth]{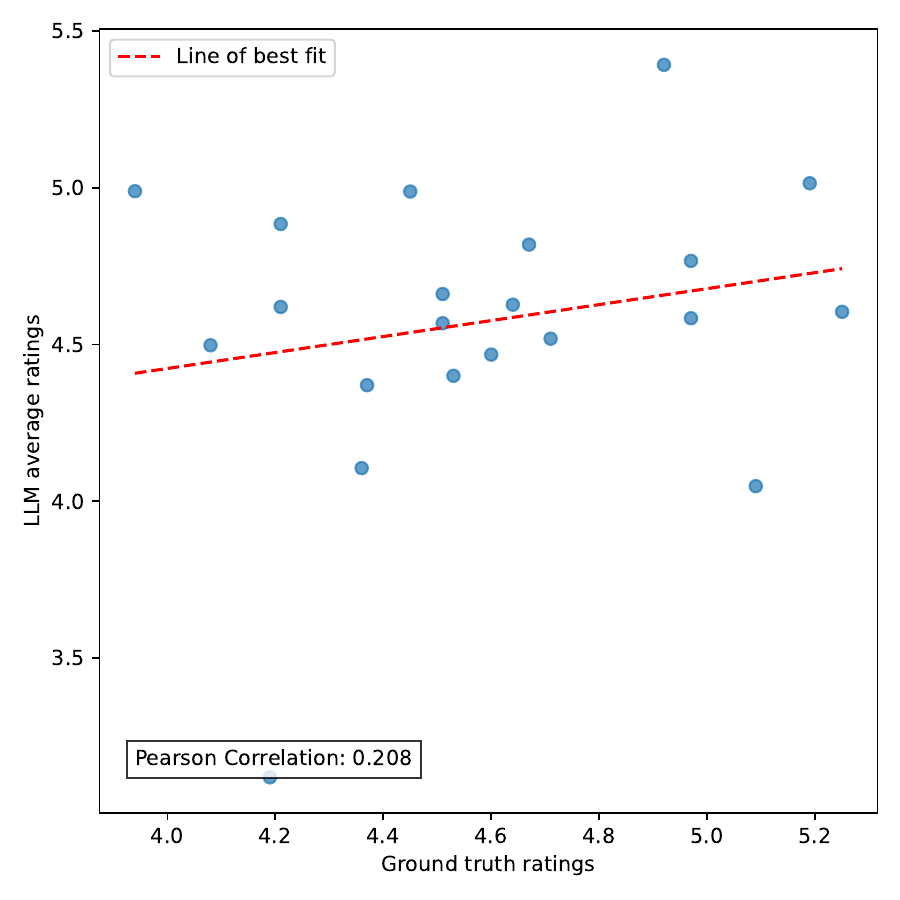}
  \caption{
  Memorization elicitation with GPT-4o: returned mean ratings versus the published Herlocker et al.~\cite{herlocker2000explaining} means, using Table~\ref{tab:memorization}. The weak correlation ($r=0.208$) indicates poor recovery under this prompt.
  }
  \label{fig:memorization}
\end{figure}

We collected 30 responses per tested method at temperature 1, averaged the returned ratings, and compared them with the published means.
The historical elicitation result was a weak Pearson correlation ($r=0.208$; Figure~\ref{fig:memorization}).
This check did not recover the published pattern reliably, indicating memorization was unlikely an issue in our tests.

\section{Exploratory audit of model-improvement responses}
\label{app:mi-case-analysis}
\paragraph{Sample and procedure.}
We reused 420 randomly selected five-attempt runs: 30 per model, with 10 from each of DCE, MCE and HYB, selected without replacement using seed 20261010. Selection did not depend on outcomes. For this four-category audit, we directly reviewed the complete parser-facing response at \emph{iteration 4} of each run, together with its supplied explanations, current configuration and previous three attempts. Thus the analysis comprises 420 labeled iterations, not 2,100 independently labeled attempts. One LLM reviewer assigned the following labels.

\paragraph{Four categories.}
\emph{Visual interpretation} records an explicit claim about the distribution image as consistent, mismatched, unassessable or absent. \emph{Tabular reasoning} flags a concrete contradiction of supplied numbers, rule conditions or configurations, rather than an unsuccessful but plausible hypothesis. \emph{Use of iterative feedback} distinguishes accurate, explicit use of earlier outcomes (G), a history/configuration contradiction or unacknowledged repeat of an ineffective proposal (Q), and no explicit outcome use (N). Labels may overlap. 

\paragraph{Improving and non-improving iterations.}
A good iteration strictly exceeds the best previously evaluated accuracy, including baseline. Of 420 fourth attempts, 40 (9.5\%) establish a new best, 66 (15.7\%) tie the previous best and 314 (74.8\%) fall below it. All have evaluated scores. We retain ties separately in the exported data; the binary association analysis groups ties and declines as non-improving. A non-improving exploratory attempt is not itself a reasoning failure.

\begin{table}[ht]
\centering\small
\caption{Iteration-4 audit: 30 runs per model (10 per condition). Feedback and tabular columns are percentages of all 30 responses; visual mismatches are relative to the retained reference image, divided by explicit visual claims (including unassessable claims). One-case changes are absolute feedback-accuracy differences of $1/116$ (approximately 0.86 percentage points) relative to the previous best.}
\resizebox{\textwidth}{!}{\begin{tabular}{lrrrrrr}
\toprule
Model & New best & Accurate feedback (\%) & Feedback problem (\%) & Tabular error (\%) & Visual mismatch/claims & One-case change\\
\midrule
Claude 4.5 Sonnet & 4/30 & 83.3 & 13.3 & 23.3 & 6/16 & 10/30 \\
Claude 4.6 Sonnet & 3/30 & 86.7 & 13.3 & 20.0 & 0/8 & 10/30 \\
Claude 4.8 Opus & 1/30 & 80.0 & 20.0 & 13.3 & 0/15 & 4/30 \\
Gemma 3 27B & 3/30 & 70.0 & 23.3 & 30.0 & 2/8 & 4/30 \\
Gemma 4 31B & 1/30 & 70.0 & 23.3 & 6.7 & 0/12 & 8/30 \\
GPT-4o & 4/30 & 46.7 & 23.3 & 30.0 & 1/15 & 2/30 \\
GPT-4o mini & 4/30 & 40.0 & 13.3 & 16.7 & 0/7 & 4/30 \\
GPT-5 & 3/30 & 93.3 & 0.0 & 6.7 & 0/10 & 7/30 \\
GPT-5 mini & 5/30 & 80.0 & 3.3 & 16.7 & 0/13 & 4/30 \\
GPT-5.5 & 3/30 & 90.0 & 6.7 & 0.0 & 0/6 & 8/30 \\
Mistral Small 3.2 & 0/30 & 46.7 & 23.3 & 16.7 & 0/1 & 4/30 \\
o4-mini & 2/30 & 43.3 & 13.3 & 40.0 & 0/1 & 8/30 \\
Qwen3-VL 32B & 4/30 & 56.7 & 36.7 & 43.3 & 1/20 & 6/30 \\
Qwen3.5 122B INT4 & 3/30 & 80.0 & 16.7 & 30.0 & 0/14 & 5/30 \\
\bottomrule
\end{tabular}}
\end{table}

\paragraph{Findings.}
Accurate explicit feedback use occurs in 290/420 responses (69.0\%); feedback problems occur in 69/420 (16.4\%), and 61/420 (14.5\%) contain no explicit use of earlier outcomes. Tabular errors occur in 88/420 (21.0\%). There are 146 explicit visual claims: 124 consistent with the retained reference, 10 mismatched and 12 unassessable. Seven mismatches arise in API responses and three in open-weight responses; the API image-provenance limitation is particularly relevant to their interpretation.

New-best proposals occur in 32/290 responses with accurate feedback use (11.0\%), versus 8/130 without this label (6.2\%). Feedback problems and tabular errors have new-best rates of 5/69 (7.2\%) and 8/88 (9.1\%), respectively. Their unlabelled counterparts have rates of 9.97\% and 9.64\%. Thus the inspected errors do not provide a strong separation between successful and unsuccessful exploration. Accurate feedback use has a small positive pooled correlation with fifth-attempt improvement over baseline, but its within-model--condition centered correlation is only 0.035 (versus 0.135 pooled). This attenuation and the uncertain association with fourth-attempt success argue against a causal interpretation or a general claim that feedback use explains performance differences.

\begin{table}[ht]\centering\small
\caption{Exploratory point-biserial correlations (phi for binary new-best outcome), with 95\% percentile intervals from 10,000 run-level resamples stratified within model and condition. Final accuracy is the evaluated fifth-attempt score (419 available; one failed attempt is missing); final gain is the percentage-point difference from the run's baseline. No multiplicity adjustment. Visual comparisons use all responses, including those without an explicit visual claim.}
\resizebox{\textwidth}{!}{\begin{tabular}{lccc}\toprule
Label & New best at attempt 4 & Attempt-5 accuracy & Attempt-5 gain over baseline\\\midrule
Accurate feedback use & 0.08 [-0.01, 0.15] & 0.06 [-0.04, 0.16] & 0.14 [0.04, 0.23] \\
Feedback problem & -0.03 [-0.11, 0.05] & 0.02 [-0.07, 0.10] & -0.04 [-0.13, 0.03] \\
Tabular error & -0.01 [-0.10, 0.09] & -0.07 [-0.17, 0.03] & -0.09 [-0.19, 0.02] \\
Visual mismatch & 0.00 [-0.06, 0.10] & 0.02 [-0.05, 0.08] & 0.01 [-0.04, 0.05] \\
\bottomrule\end{tabular}}\end{table}

Of the 40 new-best improvements, 19 (47.5\%) increase feedback accuracy by exactly $1/116$ (approximately 0.86 percentage points). Across all reviewed iterations, 84/420 (20.0\%) have an absolute feedback-accuracy difference of $1/116$ from the previous best. These counts demonstrate coarse feedback resolution, not that those improvements are spurious; repeated independent evaluation would be required to separate generalizable improvement from feedback-set optimization.


\section{Compute resources}
\label{app:compute}

Collecting new responses has different requirements: API access for the nine proprietary models, or Linux/NVIDIA hardware with sufficient GPU memory and the pinned inference stack for open-weight models.
Four open-weight checkpoints use FP8 quantisation; Qwen3.5 122B uses GPTQ INT4.
Running the entire collection took several days using a G7e GPU with 32GB RAM on AWS.
The API calls by comparison are quicker, but entail possibly significant costs.

\section{Prompt-Perturbation Robustness Experiment}
\label{app:prompt_perturbation}

\subsection{Motivation}
We conducted a supplementary experiment to assess sensitivity to minor, meaning-preserving changes in prompt wording.
This historical experiment is separate from the final 300-question teaching benchmark.

\subsection{Setup}
We used GPT-5 mini on 50 class-balanced teaching questions under the control, causal and counterfactual conditions.
Each question--condition pair received eight responses in each of two rounds, giving 1,200 calls per round and 2,400 calls in total.

\subsection{Perturbation Procedure}
\begin{itemize}
    \item \textbf{Original:} the unmodified prompt was sampled eight times.
    \item \textbf{Minor edit:} an auxiliary MLLM edited each base prompt once, replacing one to five words with synonyms or near-synonyms while preserving sentence meaning, numbers, feature names, instances and the required output format. That fixed edited prompt was then sampled eight times.
\end{itemize}

\subsection{Analysis and Results}
We compared the proportion of positive answers and the agreement of modal answers across rounds.
The question--condition pair, rather than each repeated answer, is the unit for modal agreement.
The reported historical summary gives unchanged modal answers for 148 of 150 pairs, with the two changes occurring in opposite directions.
Positive-answer rates differed by 0.4--3.3 percentage points (Table~\ref{tab:prompt-robustness}).
The response-level two-proportion tests shown below are exploratory summaries and do not account for dependence among answers to the same question.

\begin{table}[ht]
\centering\small
\begin{tabular}{lccccc}
\toprule
Condition & \multicolumn{2}{c}{Positive rate} & Modal & Flips & $p$\\
\cmidrule(lr){2-3}
 & Original & Edited & agreement & (of 50) & (2-prop.)\\
\midrule
Control & 0.421 & 0.417 & 1.00 & 0 & 0.918\\
Causal & 0.409 & 0.376 & 0.98 & 1 & 0.341\\
Counterfactual & 0.495 & 0.472 & 0.98 & 1 & 0.520\\
\midrule
All & 0.441 & 0.422 & --- & 2 & 0.335\\
\bottomrule
\end{tabular}
\caption{Historical prompt-perturbation summary: 50 questions, three conditions, eight responses per round, and GPT-5 mini. Positive rate is the proportion of answers equal to 1. Modal agreement compares aggregated answers between rounds. The reported response-level $p$-values are unadjusted exploratory summaries.}
\label{tab:prompt-robustness}
\end{table}

\section{Mental-model alignment through recourse rationales}
\label{app:recourse-rationales}

We investigated mental-model alignment through the rationales people and MLLMs express about classifier rejection and proposed changes. Specifically, do similar expressed beliefs about justification, feasibility and burden accompany similar ratings? Tominaga et al.~\cite{tominaga2024reassessing} report 13 human quotations linked to acceptance or willingness ratings; all are listed in Table~\ref{tab:recourse-rationale-themes}, grouped into eight themes. These illustrative cases motivate directional expectations, rather than establish population-level human correlations. Our comparison concerns expressed understanding, not direct access to internal mental models.

We sampled 30 existing responses per model using a seed-42 shuffled order (420 responses across 14 models). One automated labeller (GPT-6 Astra) applied a fixed codebook with model identities and explicit score fields masked. For each model and theme, we correlated label presence with its designated rating using Spearman's $\rho$, excluding uncertain labels. We multiplied $\rho$ by the expected direction so positive values indicate alignment. We obtained 95\% percentile intervals from 10,000 bootstrap resamples of applicant clusters.

\begin{figure}[htbp]
\centering
\includegraphics[width=\linewidth]{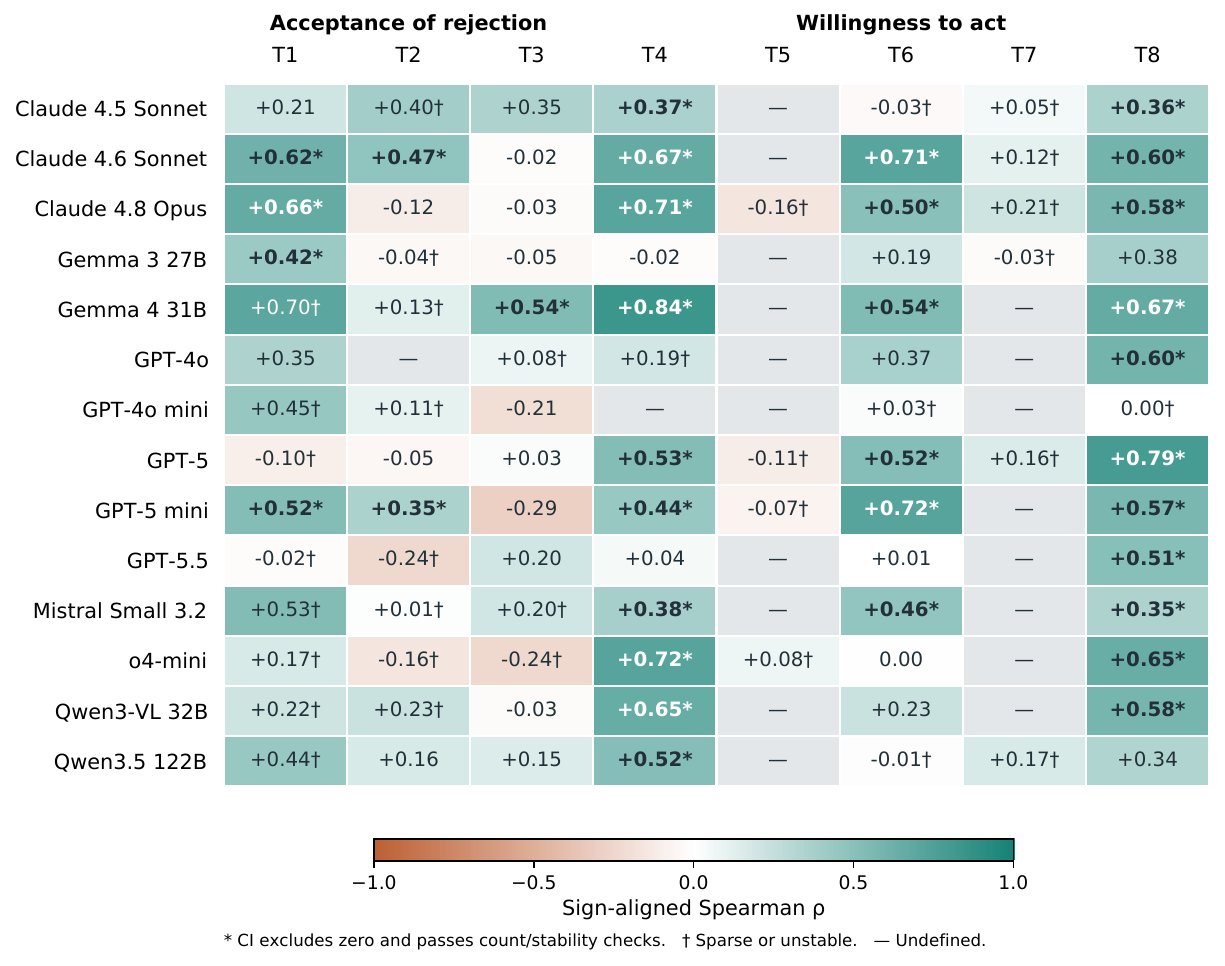}
\caption{Associations between expressed recourse rationales and ratings, separately for each model. Positive sign-aligned $\rho$ follows the human-example direction in Table~\ref{tab:recourse-rationale-themes}. Bold starred cells have 95\% intervals excluding zero, at least five label-present and five label-absent cases, and at least 95\% defined bootstrap replicates. Daggers flag sparse or unstable estimates; dashes denote undefined correlations. Applicant clustering accounts for multiple plans for the same applicant. Rare labels cannot be assessed reliably from this pilot.}
\label{fig:recourse-rationale-heatmap}
\end{figure}

Feasibility (T4) accompanied higher willingness, and disproportionate effort (T8) accompanied lower willingness: 10/14 and 11/14 models, respectively, had intervals entirely in the expected direction and passed the count/stability checks in Figure~\ref{fig:recourse-rationale-heatmap}. Small-change criticism (T1) had the expected point-estimate direction in 12/14 models, but only four passed these interval and count checks. Other themes were mixed or rare.

\begin{table}[htbp]
\centering
\small
\setlength{\tabcolsep}{3pt}
\renewcommand{\arraystretch}{1.12}
\caption{All 13 human quotations used in the audit, grouped into eight themes, from Tominaga et al.~\cite{tominaga2024reassessing}, pp.~6--7. A denotes acceptance of the original rejection; W denotes willingness to act. Ratings are out of seven. Arrows give the expected association with theme presence. Short quotations are interpreted using the authors' surrounding text.}
\label{tab:recourse-rationale-themes}
\begin{tabularx}{\linewidth}{@{}l p{.22\linewidth} X l@{}}
\toprule
Code & Theme & Human quotation (participant) & Rating; direction \\
\midrule
T1 & Small-change criticism & ``Almost same'' (P006) & A: 1; $\downarrow$ \\
T1 & & ``That's it?'' (P446) & A: 2; $\downarrow$ \\
T1 & & ``Not particularly different compared to the current situation'' (P706) & A: 2; $\downarrow$ \\
T2 & Large gap justifies rejection & ``The career is too different'' (P567) & A: 7; $\uparrow$ \\
T2 & & ``I think it's a solid reason, so I don't blame it'' (P739) & A: 6; $\uparrow$ \\
T3 & Personal shortcoming & ``Because I realize how much I'm not enough'' (P732) & A: 5; $\uparrow$ \\
T4 & Readily feasible change & ``Because I think I can tackle it right now'' (P674) & W: 5; $\uparrow$ \\
T4 & & ``If it was just working hours, I'd do it in a heartbeat'' (P718) & W: 7; $\uparrow$ \\
T4 & & ``Because the contents are easy to change'' (P294) & W: 7; $\uparrow$ \\
T5 & Waiting is sufficient & ``All I have to do is wait a little'' (P655) & W: 6; $\uparrow$ \\
T6 & Time or money burden & ``Because there's too much to do, it takes too much time and money'' (P401) & W: 2; $\downarrow$ \\
T7 & Too many requirements & ``There are too many items to clear'' (P167) & W: 1; $\downarrow$ \\
T8 & Disproportionate effort & ``I don't want to change various things just to apply for a loan'' (P468) & W: 1; $\downarrow$ \\
\bottomrule
\end{tabularx}
\end{table}

\section{Correlation between human mean differences and ranking success}
\label{app:human_means_vs_rankings}
\begin{figure}[!ht]
\centering
  \includegraphics[width=0.8\textwidth]{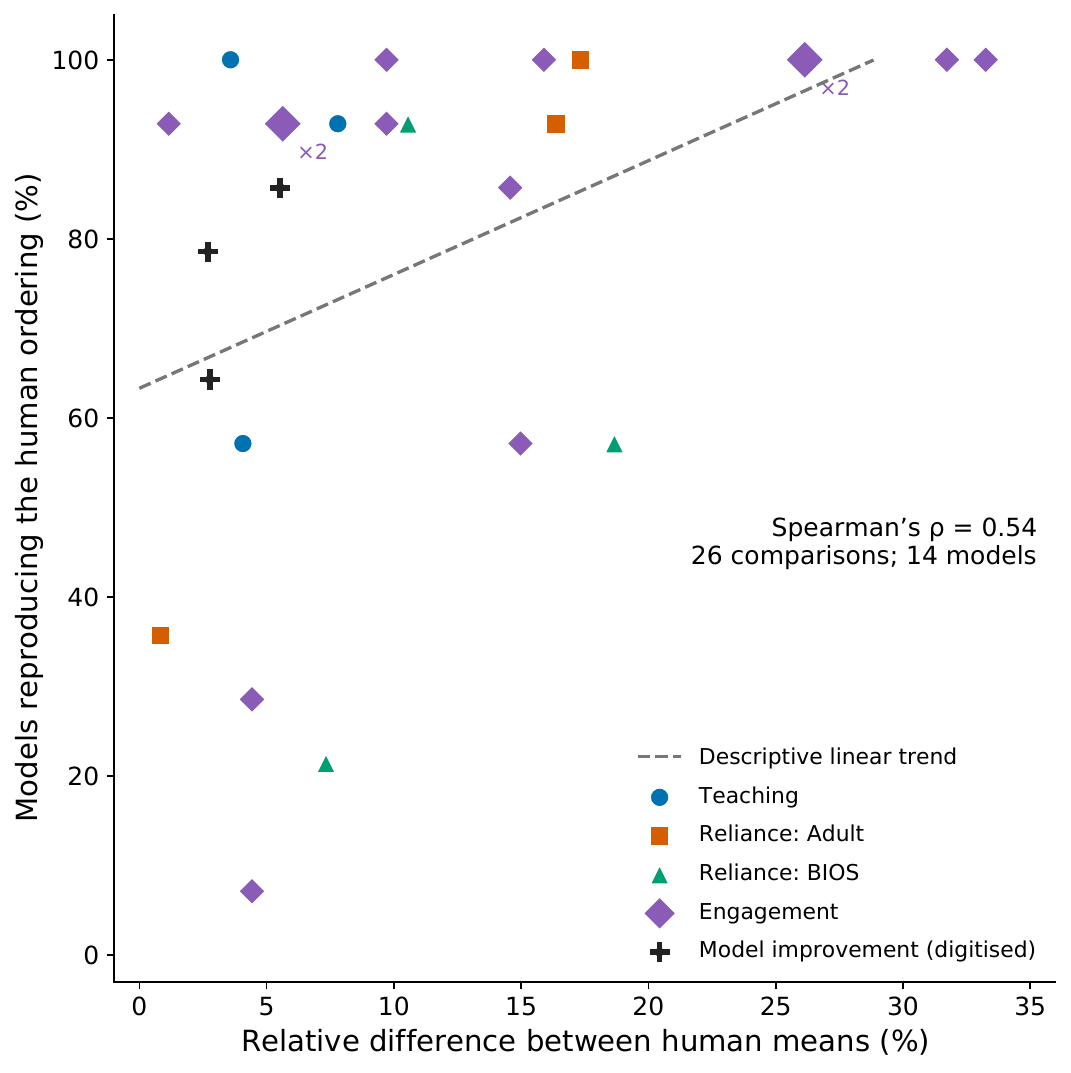}
  \caption{
  Human performance differences and MLLM ranking agreement. Each point represents one of 26 condition comparisons across four human studies (no recourse study). The x-axis shows the relative difference between human means, \(100(\mu_{\mathrm{higher}}-\mu_{\mathrm{lower}})/\mu_{\mathrm{lower}}\); the y-axis shows the percentage of 14 MLLMs reproducing that ordering. Colours distinguish study domains; black crosses indicate Model Improvement comparisons using means estimated from published participant dots. Larger markers labelled ×2 represent overlapping comparisons. The dashed line shows a descriptive linear trend (Spearman’s \(\rho=0.54\)). The empty lower-right region suggests fewer ranking failures for large human differences within these studies. Comparisons share conditions and are not independent.
  }
  \label{fig:human_vs_rankings}
\end{figure}

We examined whether larger differences between human study outcomes were associated with more consistent reproduction of their rankings by MLLMs. For each of 26 condition comparisons across four studies, we calculated the relative difference between published human means, estimating Model Improvement means by digitising the participant scores in its published figure. We plotted this difference against the percentage of 14 MLLMs reproducing the same ordering. Larger human differences were associated with greater agreement (Spearman’s $\rho=0.54$). The absence of points in the lower-right region of Figure~\ref{fig:human_vs_rankings} indicates that large human differences were rarely accompanied by widespread ranking failures among the tested models. This descriptive pattern suggests that larger human differences may be easier to reproduce.

The comparisons used for Figure~\ref{fig:human_vs_rankings} are in Table~\ref{tab:human-gap-comparisons}.

\begin{table*}[t]
\centering
\caption{The 26 comparisons included in the human-gap analysis.
The symbol $>$ denotes a higher human mean, not necessarily a
statistically significant difference. NN denotes nearest neighbours;
E1--E21 identify engagement explanation conditions. E4 versus E5 is
excluded because their reported means are equal. Model Improvement
means were estimated from participant points in the published figure.
Recourse is excluded because comparable discrete condition means
are unavailable.}
\label{tab:human-gap-comparisons}
\resizebox{\linewidth}{!}{%
\begin{tabular}{ll@{\hspace{1cm}}ll}
\toprule
Study/domain & Human ordering & Study/domain & Human ordering \\
\midrule
Teaching & Counterfactual $>$ Control
  & Engagement & E1 $>$ E21 \\
Teaching & Counterfactual $>$ Causal
  & Engagement & E2 $>$ E4 \\
Teaching & Causal $>$ Control
  & Engagement & E2 $>$ E5 \\
Reliance---Adult & NN $>$ Control
  & Engagement & E2 $>$ E11 \\
Reliance---Adult & NN $>$ Saliency
  & Engagement & E2 $>$ E21 \\
Reliance---Adult & Control $>$ Saliency
  & Engagement & E4 $>$ E11 \\
Reliance---BIOS & NN $>$ Control
  & Engagement & E4 $>$ E21 \\
Reliance---BIOS & NN $>$ Saliency
  & Engagement & E5 $>$ E11 \\
Reliance---BIOS & Saliency $>$ Control
  & Engagement & E5 $>$ E21 \\
Engagement & E1 $>$ E2
  & Engagement & E11 $>$ E21 \\
Engagement & E1 $>$ E4
  & Model Improvement & HYB $>$ DCE \\
Engagement & E1 $>$ E5
  & Model Improvement & HYB $>$ MCE \\
Engagement & E1 $>$ E11
  & Model Improvement & DCE $>$ MCE \\
\bottomrule
\end{tabular}%
}
\end{table*}

\newpage

\begin{figure}[!ht]
\centering
  \includegraphics[width=\textwidth]{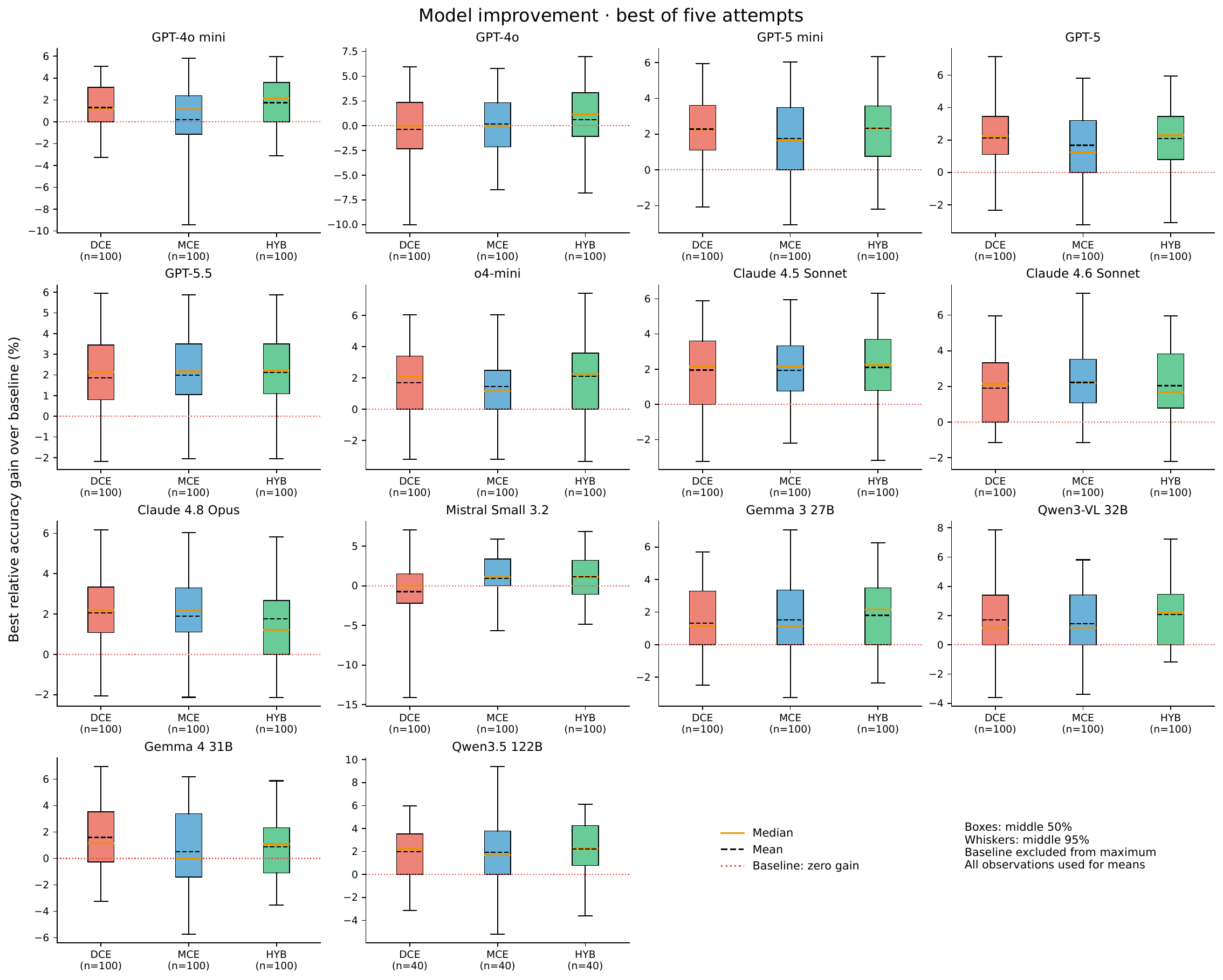}
  \caption{All-model model-improvement results: best successfully evaluated proposal among five attempts, excluding baseline from the maximum, expressed as relative percentage improvement over the run baseline. There are 100 runs per condition, except 40 for Qwen3.5 122B. Boxes show the interquartile range and median; whiskers span the central 95\% of observations. Outliers omitted from display still contribute to the mean.}
  \label{fig:model_impro_all_results}
\end{figure}

\newpage

\begin{figure}[!ht]
\centering
  \includegraphics[width=\textwidth]{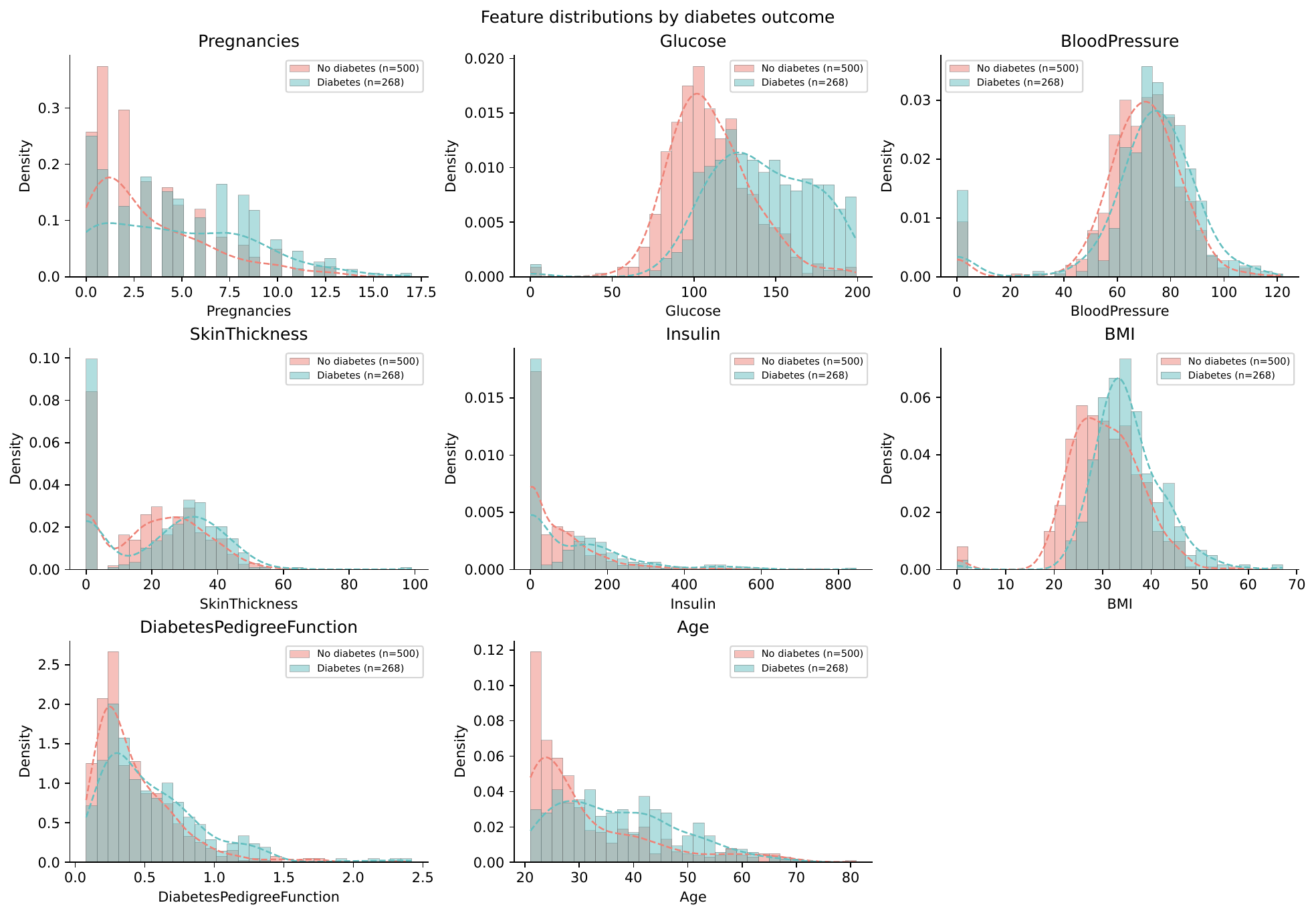}
  \caption{Publication rendering of the feature distributions used in model improvement. The original static image supplied to the open-weight DCE and HYB conditions is preserved separately in the release; this grid-free rendering is not a replacement for the collection input.}
  \label{fig:model_improv_distributions}
\end{figure}

\newpage

\begin{figure}[!ht]
\centering
  \includegraphics[width=\textwidth]{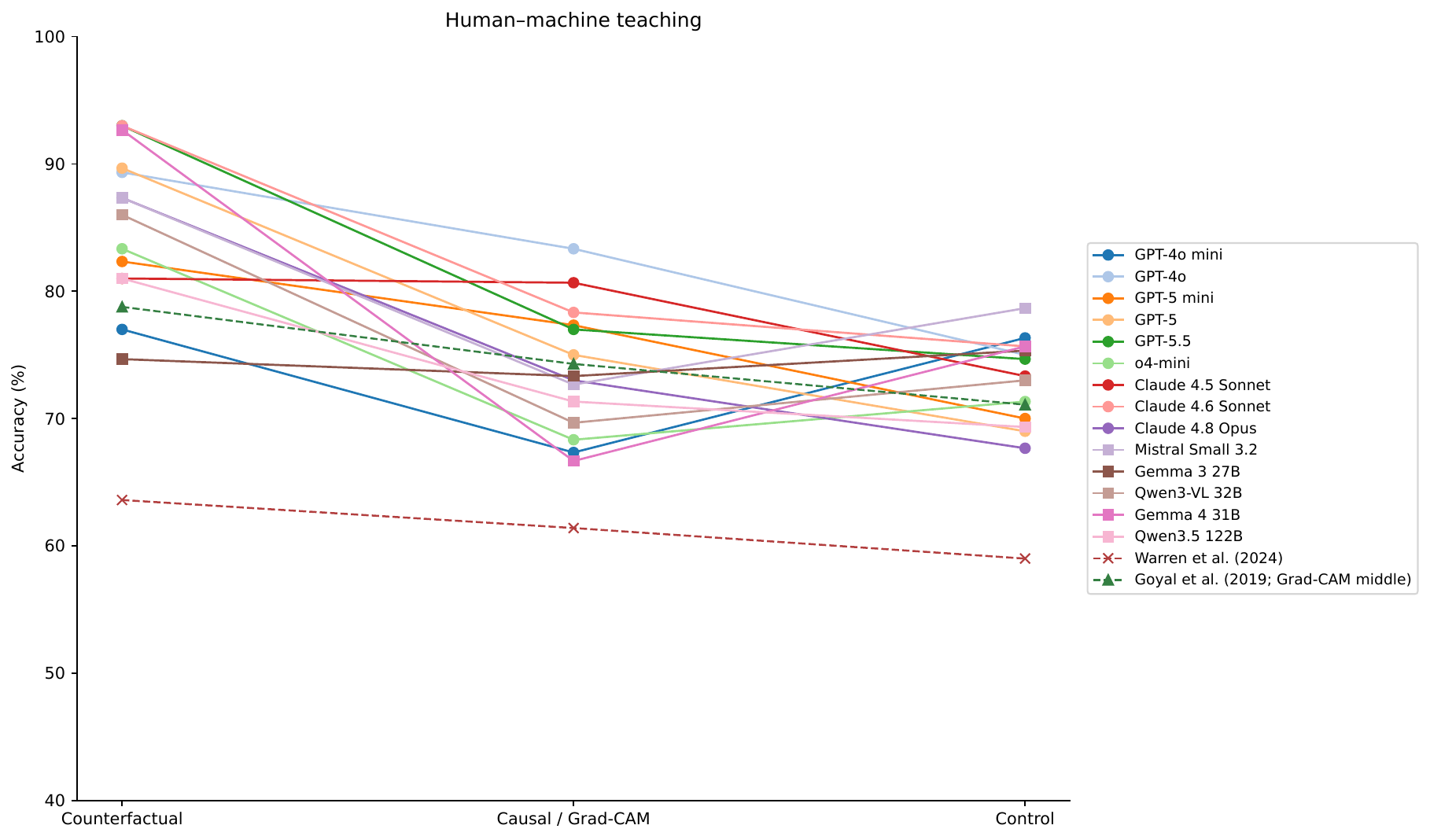}
  \caption{All-model teaching accuracy over the 300 matched synthetic questions per condition, compared descriptively with published human means. Warren et al.\ provide the benchmark\textquotesingle s teaching reference; Goyal et al.\ provide a separate comparison whose middle condition is Grad-CAM rather than causal explanations. Invalid model answers count as incorrect.}
  \label{fig:human_mac_teaching_all_results}
\end{figure}

\newpage

\begin{figure}[!ht]
\centering
  \includegraphics[width=\textwidth]{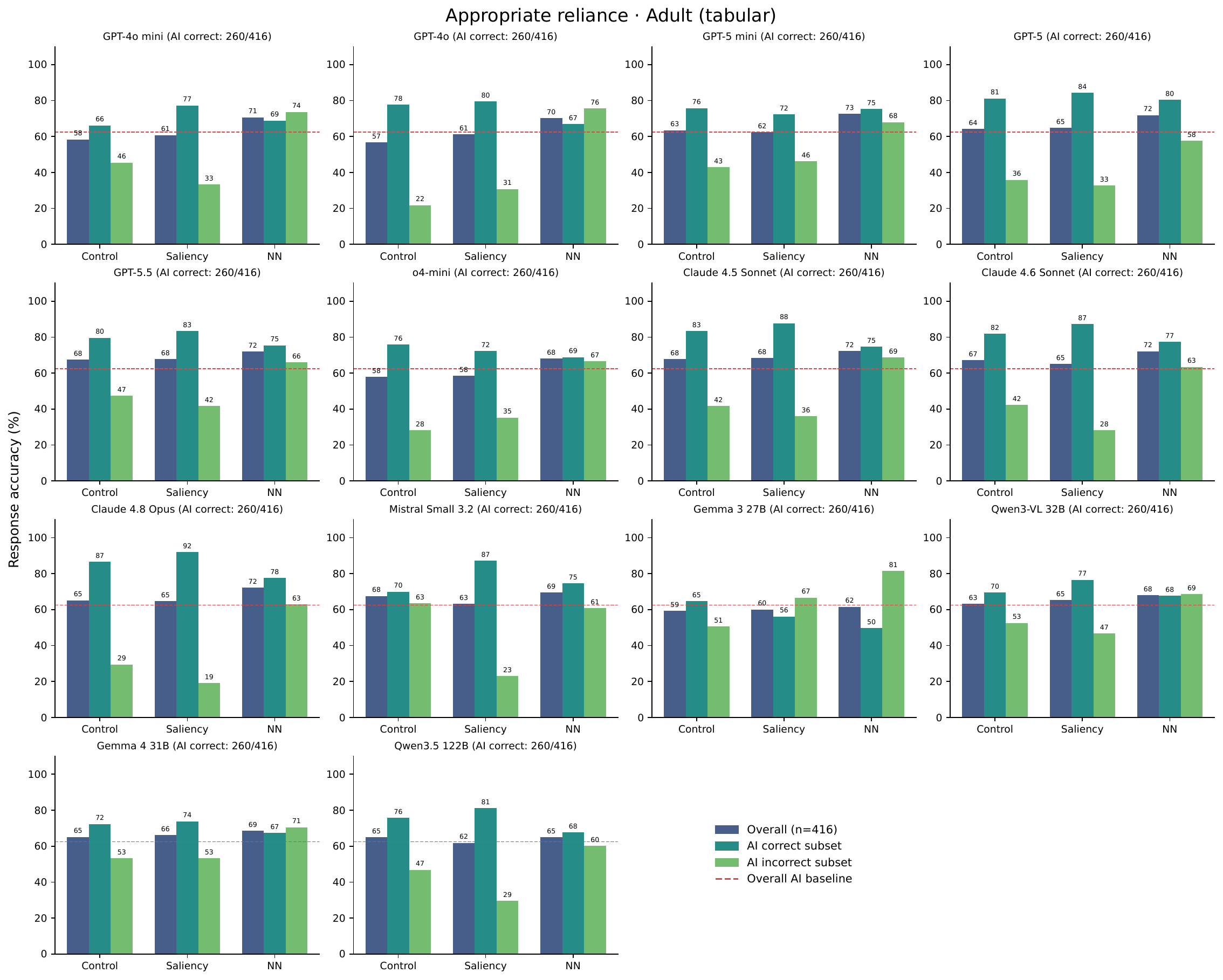}
  \caption{Adult Census reliance results over 416 cases per condition. Overall denotes classification accuracy; AI Correct and AI Incorrect restrict cases by the classifier\textquotesingle s actual correctness. AI Baseline is classifier accuracy on the selected cases (260/416). Control has no AI advice; the saliency and nearest-neighbour conditions include advice. These bars do not establish significance of condition differences.}
  \label{fig:reliance_all_results_adult}
\end{figure}

\newpage

\begin{figure}[!ht]
\centering
  \includegraphics[width=\textwidth]{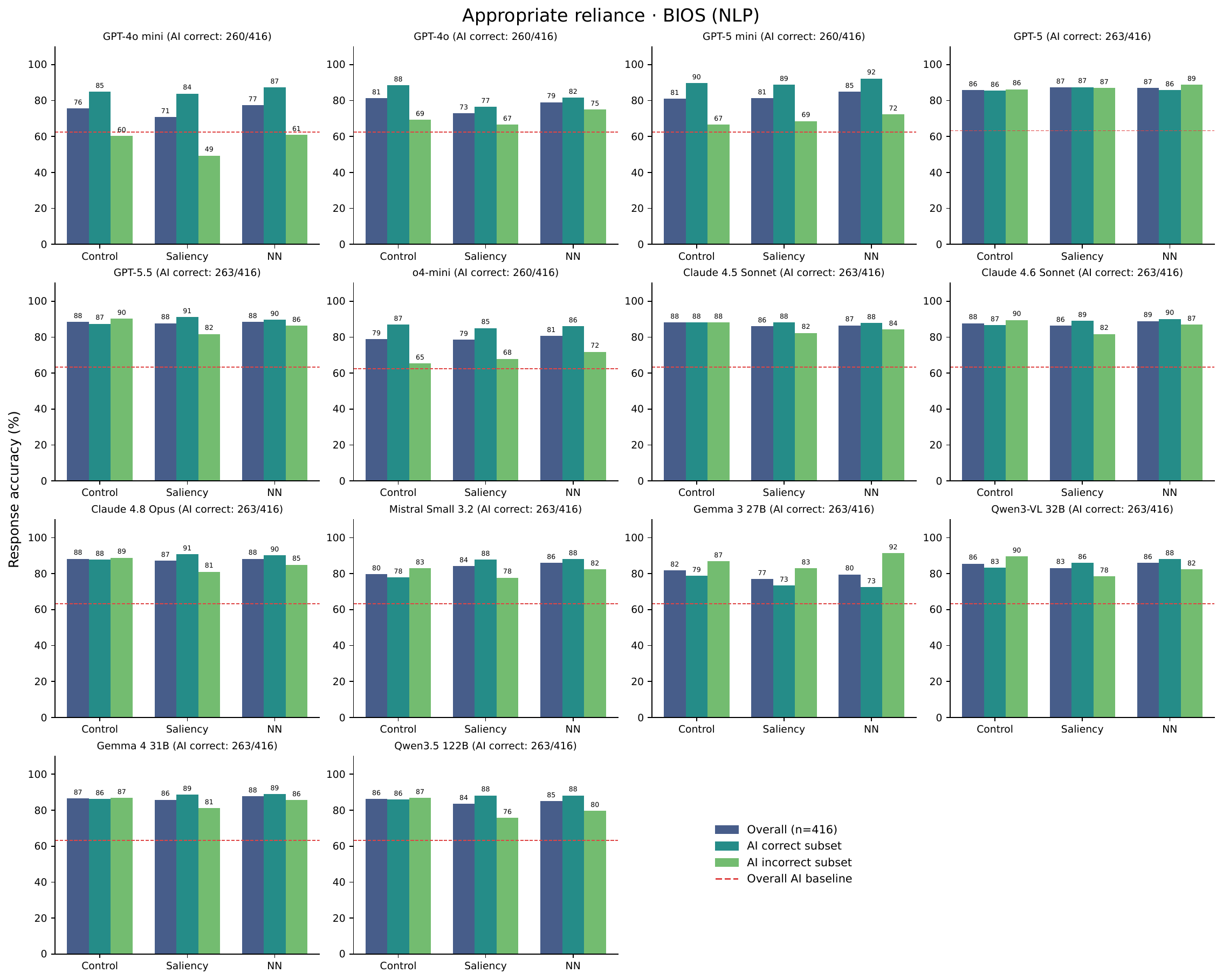}
  \caption{BIOS reliance results over 416 cases per condition. Overall denotes classification accuracy; AI Correct and AI Incorrect restrict cases by the classifier\textquotesingle s actual correctness. AI Baseline is 260/416 for GPT-4o, GPT-4o mini, GPT-5 mini and o4-mini, and 263/416 for the other models. Some API case sets differ, so cross-model bars are descriptive rather than paired case comparisons.}
  \label{fig:reliance_all_results_bios}
\end{figure}

\newpage

\newpage

\begin{figure}[!ht]
\centering
  \includegraphics[width=\textwidth]{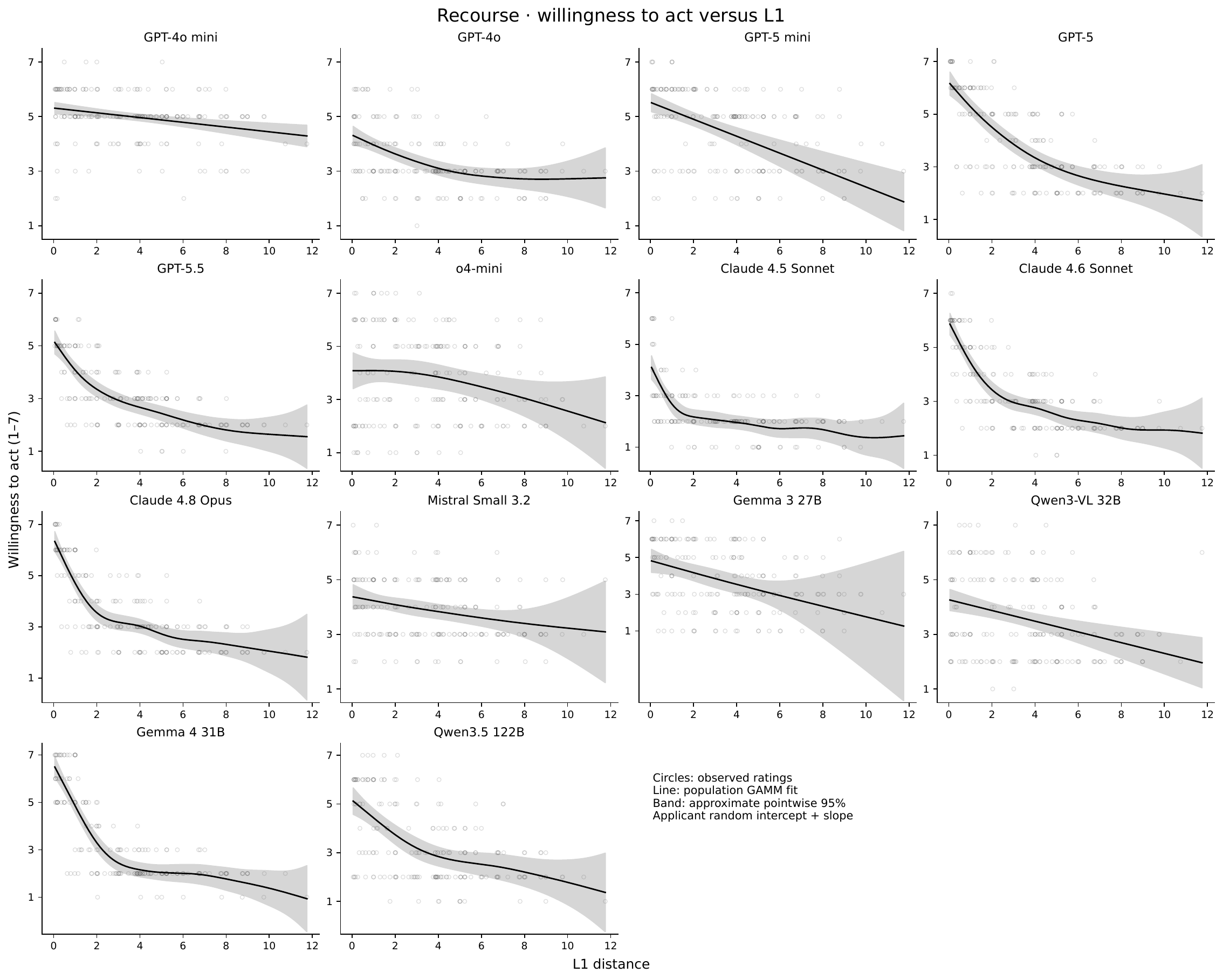}
  \caption{Willingness to act versus mixed $L_1$ distance: training-standard-deviation-scaled numeric changes plus one unit per changed categorical field. Points show valid ratings; curves are population-mean GAMM trends, with approximate pointwise 95\% mean intervals. The 200 plans share 39 applicants; the relationship is not a matched human-rating comparison.}
  \label{fig:actionable_rec_all_results2}
\end{figure}

\newpage

\begin{figure}[!ht]
\centering
  \includegraphics[width=\textwidth]{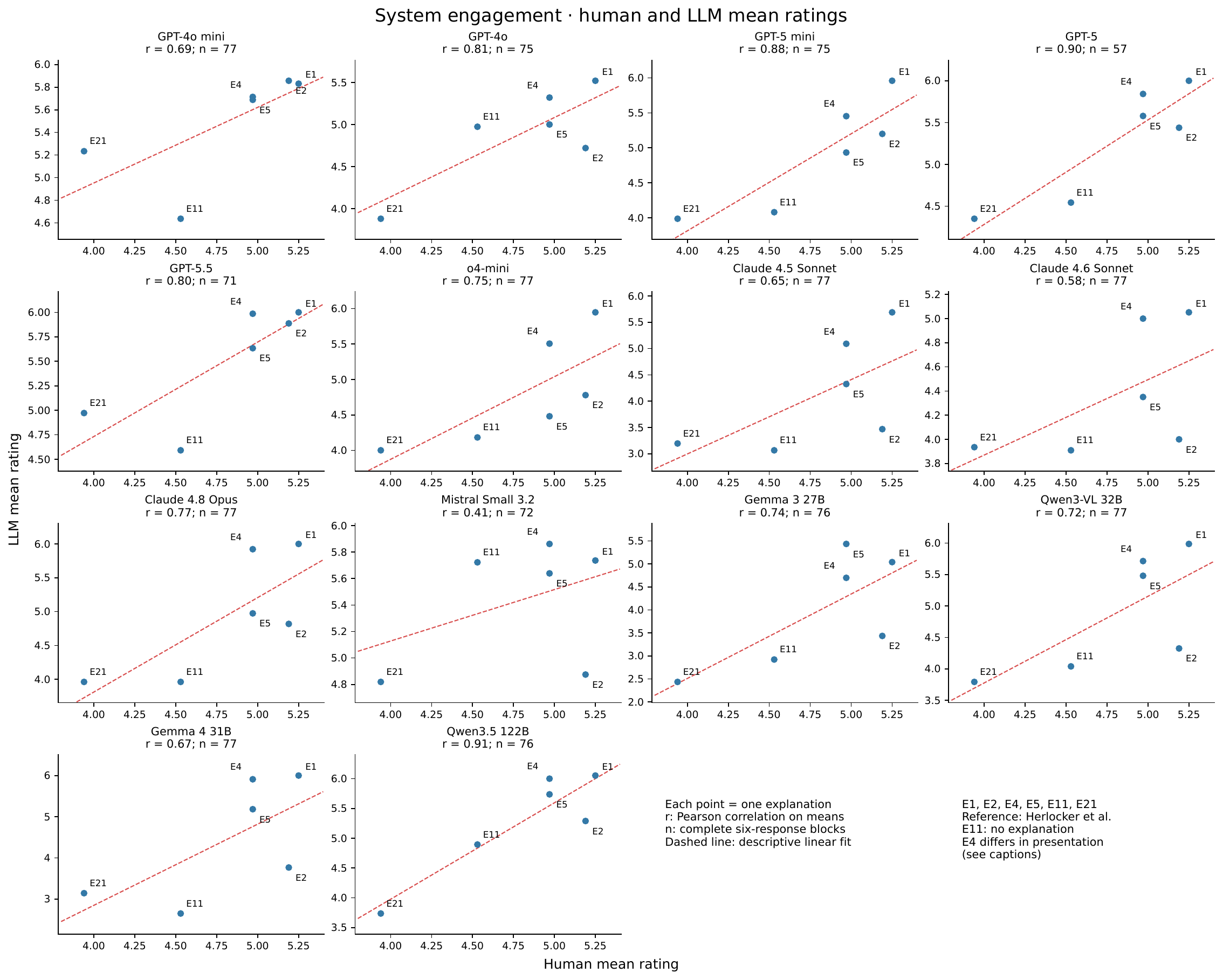}
  \caption{System-engagement comparisons between the six published human condition means and model means from complete response blocks. Each panel reports Pearson\textquotesingle s $r$ over six means; its response-block count is not the correlation sample size. The primary table uses Spearman\textquotesingle s $\rho$. Panel scales can differ.}
  \label{fig:sys_eng_all_results}
\end{figure}

\clearpage

\section*{NeurIPS Paper Checklist}

\providecommand{\answerYes}{\textbf{Yes}}
\providecommand{\answerNo}{\textbf{No}}
\providecommand{\answerNA}{\textbf{NA}}

\begin{enumerate}
\item \textbf{Claims}
\item[] Question: Do the main claims made in the abstract and introduction accurately reflect the paper's contributions and scope?
\item[] Answer: \answerYes{}.
\item[] Justification: The paper evaluates 14 models across five purpose-grounded studies and reports model- and task-dependent agreement. Its conceptual analysis concerns sufficient conditions for rank preservation; the empirical results support preliminary evaluation and prioritising human studies.

\item \textbf{Limitations}
\item[] Question: Does the paper discuss the limitations of the work performed by the authors?
\item[] Answer: \answerYes{}.
\item[] Justification: We describe reconstructed study materials, fixed human references, conditional bootstrap uncertainty, feedback-set optimisation, model/version dependence, and the limited scope of contamination and prompt-perturbation checks. These results do not establish population-level human validity.

\item \textbf{Theory assumptions and proofs}
\item[] Question: For each theoretical result, does the paper provide the full set of assumptions and a complete (and correct) proof?
\item[] Answer: \answerYes{}.
\item[] Justification: Section~\ref{sec:theory} defines condition-specific calibration error. For a positive human mean gap $\Delta_H$, the corresponding model gap is $\Delta_H+\epsilon_i-\epsilon_j$, which is positive whenever $|\epsilon_i-\epsilon_j|<\Delta_H$. This is a sufficient algebraic condition, not an empirical guarantee that the assumptions hold.

\item \textbf{Experimental result reproducibility}
\item[] Question: Does the paper fully disclose all the information needed to reproduce the main experimental results of the paper to the extent that it affects the main claims and/or conclusions of the paper (regardless of whether the code and data are provided or not)?
\item[] Answer: \answerYes{}.
\item[] Justification: The release supplies frozen analysis inputs, available prompts/responses, manifests and scripts for reproducing the primary table and figures. Missing historical collection provenance is identified. Reproducing analysis from saved data is distinguished from re-running stochastic inference.

\item \textbf{Open access to data and code}
\item[] Question: Does the paper provide open access to the data and code, with sufficient instructions to faithfully reproduce the main experimental results, as described in supplemental material?
\item[] Answer: \answerYes{}.
\item[] Justification: The repository provides the analysis inputs, code, pinned dependencies and reproduction instructions. Third-party data and study materials retain their applicable terms.

\item \textbf{Experimental setting/details}
\item[] Question: Does the paper specify all the training and test details (e.g., data splits, hyperparameters, how they were chosen, type of optimizer, etc.) necessary to understand the results?
\item[] Answer: \answerYes{}.
\item[] Justification: The study descriptions specify task inputs, conditions, sampling units, available model settings, classifier configurations and endpoints. The release records additional settings and provenance limits.

\item \textbf{Experiment statistical significance}
\item[] Question: Does the paper report error bars suitably and correctly defined or other appropriate information about the statistical significance of the experiments?
\item[] Answer: \answerYes{}.
\item[] Justification: The primary table reports full-data Spearman estimates and 95\% percentile intervals from 10,000 matched bootstrap resamples at observed sample sizes. Human references are fixed for the condition-ranking studies; recourse instead correlates negative $L_1$ cost with willingness, conditional on the observed applicant pool. Figure captions distinguish descriptive distributions, standard errors and GAMM population-mean bands. A nonnegative lower interval bound is a formatting convention rather than evidence of strictly positive correlation.

\item \textbf{Experiments compute resources}
\item[] Question: For each experiment, does the paper provide sufficient information on the computer resources (type of compute workers, memory, time of execution) needed to reproduce the experiments?
\item[] Answer: \answerYes{}
\item[] Justification: Appendix~\ref{app:compute} details it.

\item \textbf{Code of ethics}
\item[] Question: Does the research conducted in the paper conform, in every respect, with the NeurIPS Code of Ethics \url{https://neurips.cc/public/EthicsGuidelines}?
\item[] Answer: \answerYes{}.
\item[] Justification: We use existing study references and model-generated responses, preserve source attribution, and remove credentials and private deployment identifiers from the release.

\item \textbf{Broader impacts}
\item[] Question: Does the paper discuss both potential positive societal impacts and negative societal impacts of the work performed?
\item[] Answer: \answerYes{}.
\item[] Justification: Surrogate evaluation can help prioritise human studies and reduce reliance on poorly targeted proxy metrics. Misusing model agreement as a substitute for human evidence could favour explanations that perform well for models but poorly for affected people. Our recommendations retain human validation and reporting of uncertainty.

\item \textbf{Safeguards}
\item[] Question: Does the paper describe safeguards that have been put in place for responsible release of data or models that have a high risk for misuse (e.g., pretrained language models, image generators, or scraped datasets)?
\item[] Answer: \answerNA{}.
\item[] Justification: We do not release model weights or a new high-risk generative model. The released benchmark preserves dataset restrictions and redacts credentials and private infrastructure identifiers; generated text is not presented as human participant data.

\item \textbf{Licenses for existing assets}
\item[] Question: Are the creators or original owners of assets (e.g., code, data, models), used in the paper, properly credited and are the license and terms of use explicitly mentioned and properly respected?
\item[] Answer: \answerYes{}.
\item[] Justification: We cite the source studies and datasets and provide checkpoint and license notices.

\item \textbf{New assets}
\item[] Question: Are new assets introduced in the paper well documented and is the documentation provided alongside the assets?
\item[] Answer: \answerYes{}.
\item[] Justification: The release documents synthetic teaching inputs, reconstructed task materials, saved model responses, analysis code, schemas, provenance, licenses and reproducibility limits. Project-owned data and annotations are offered under CC BY 4.0 with explicit third-party exclusions.

\item \textbf{Crowdsourcing and research with human subjects}
\item[] Question: For crowdsourcing experiments and research with human subjects, does the paper include the full text of instructions given to participants and screenshots, if applicable, as well as details about compensation (if any)?
\item[] Answer: \answerNA{}.
\item[] Justification: We did not recruit new human participants. Human findings are taken from the cited studies, and our new responses are generated by MLLMs.

\item \textbf{Institutional review board (IRB) approvals or equivalent for research with human subjects}
\item[] Question: Does the paper describe potential risks incurred by study participants, whether such risks were disclosed to the subjects, and whether Institutional Review Board (IRB) approvals (or an equivalent approval/review based on the requirements of your country or institution) were obtained?
\item[] Answer: \answerNA{}.
\item[] Justification: No new human-subject experiment was conducted. We report published human-study findings and generate new model responses.

\item \textbf{Declaration of MLLM usage}
\item[] Question: Does the paper describe the usage of MLLMs if it is an important, original, or non-standard component of the core methods in this research? Note that if the MLLM is used only for writing, editing, or formatting purposes and does not impact the core methodology, scientific rigorousness, or originality of the research, declaration is not required.
\item[] Answer: \answerYes{}.
\item[] Justification: MLLM surrogate evaluation is the core method. We describe the models, prompts, response processing, sampling and analysis throughout the paper and appendix.
\end{enumerate}

\end{document}